\pdfoutput=1

\documentclass[11pt]{article}

\usepackage[final]{acl}

\usepackage[many]{tcolorbox}
\tcbuselibrary{breakable, skins}
\usepackage{fvextra} 
\usepackage{caption} 
\usepackage[normalem]{ulem}

\usepackage{times}
\usepackage{latexsym}
\usepackage{url}
\usepackage{pgfplots}
\usepackage{booktabs}
\usepackage{siunitx}
\usepackage{makecell}
\usepackage{caption}
\usepackage{subcaption}
\usepackage{tabularray}
\usepackage{amssymb}
\usepackage{amsmath}
\usepackage{enumitem}
\usepackage{multirow}

\newcounter{prompt}

\usepackage[T1]{fontenc}

\usepackage[utf8]{inputenc}

\usepackage{microtype}

\usepackage{inconsolata}

\usepackage{graphicx}

\title{OARelatedWork: A Large-Scale Dataset of Related Work Sections with Full-texts from Open Access Sources}

\author{
  Martin Docekal, Martin Fajcik, Pavel Smrz
  \\
  Brno University of Technology
  \\
  \texttt{\{idocekal, ifajcik, smrz\}@fit.vut.cz}
}

\begin{document}
\maketitle
\begin{abstract}

  This paper introduces OARelatedWork: a dataset for related work generation from open-access sources.\footnote{Dataset: \url{https://huggingface.co/datasets/BUT-FIT/OARelatedWork}
  
  Meta-Evaluation Dataset: \url{https://huggingface.co/datasets/BUT-FIT/OARelatedWorkMetaEval}
  
  Main Repository: \url{https://github.com/KNOT-FIT-BUT/OAPapers}.} It is the first large-scale multi-document summarization dataset for related work generation, containing whole related work sections and full texts of cited papers. Its validation and test splits are constructed so that every source paper is available in full text, enabling controlled evaluation of full-text related work generation. The dataset includes 94\,450~papers from multiple domains. With OARelatedWork, we aim to shift the field from generating parts of related work sections from abstracts only to generating entire related work sections from all available content. We (i) benchmark a wide spectrum of models, highlighting that synthesizing massive full-text contexts remains challenge  for tested cost-effective Large Language Models (LLMs): under our statement-level judge, GPT-4o-mini's evidence-grounded \textit{True} rate drops from 92.9\% with abstracts to 83.8\% with full texts. We (ii) empirically analyze human writing behavior through a human evaluation over 40 papers and 408 factual statements, revealing that authors frequently introduce abstractive claims ungrounded in localized source texts; consequently, tested LLMs actually surpass human baselines in strict, evidence-grounded factuality. Finally, we (iii) conduct a fine-grained meta-evaluation, revealing that standard reference-based metrics are inadequate for evaluating such long-form structured outputs, and introduce a robust statement-level evaluation framework to address this gap.

\end{abstract}

\section{Introduction}

Related work generation extends multi-document summarization by not only summarizing referenced papers but also positioning a work within prior research through comparisons and contrasts.
Automating this process holds immense practical value: it aids researchers by drafting comparative summaries, significantly reducing the time spent on literature exploration. Furthermore, it pushes NLP models beyond simple summarization to handle complex, discourse-level generation, serving as a highly challenging benchmark for long-context reasoning.

\begin{figure}
    \centering
    \includegraphics[width=5cm]{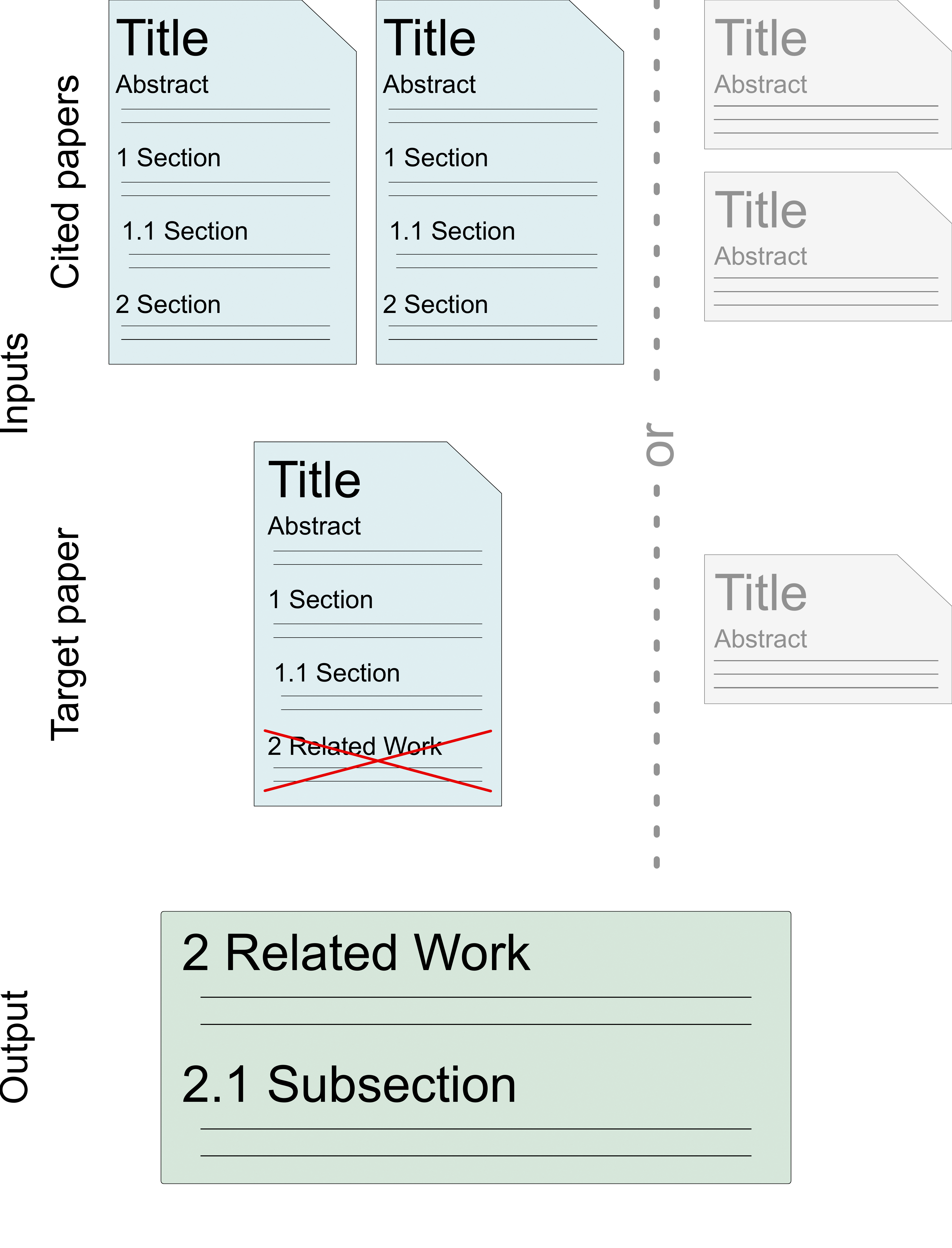}
    \caption{The task is to generate a whole related work section from cited papers and the rest of the target paper. We also try variants using abstracts instead of full-texts.}
    \label{fig:task_ilustration}
\end{figure}

As shown in Figure~\ref{fig:task_ilustration}, related work generation is demanding due to extremely long input contexts, as models must process the target paper together with all cited references, while datasets must also provide access to all required papers, many of which may be paywalled.
Furthermore, even when they are available, it is hard to automatically process them, as they are usually in PDF format. Fortunately, there are existing tools for conversion to more machine-friendly formats. For example, GROBID~\cite{GROBID}, which was specifically designed to process scientific papers and corresponds to the state-of-the-art in its field~\cite{Meuschke2023ABO}.

Although the pioneering work of \citet{Hoang2010TowardsAR} relied on full-text references and a topic hierarchy tree for a limited dataset, nowadays only abstracts are usually used~\cite{Lu2020MultiXScienceAL,Chen2021CapturingRB,Chen2022TargetawareAR,Liu2023CausalIF,structured_related_work}. This shift came probably with the raise of abstractive summarization models for which long inputs are technically demanding, given the constraints of existing models. Furthermore, the task of related work generation is commonly limited~\cite{Lu2020MultiXScienceAL, Chen2021CapturingRB} to the generation of separate paragraphs. This makes the task easier, but also, such a simplification makes it impossible to assess whether a model is able to generate a well-structured related work section, as it only benchmarks models on a short sub-summary.

Generating an entire related work section is challenging not only for the generator but also for the evaluation. While automatic evaluation of summarization remains an open problem~\cite{Fabbri2020SummEvalRS}, the challenges are even more pronounced in the context of long-form summarization.

This work addresses the issues mentioned by:
\begin{itemize}
    \item We introduce OARelatedWork (Section~\ref{sec:dataset}), providing automatically processed full texts of cited and target papers to benchmark the generation of entire related work sections. Through this dataset, we empirically analyze human-authored related work sections and compare them with generated ones.

    \item We evaluate diverse baselines (Section~\ref{sec:results}) and conduct a human evaluation (Section~\ref{sec:human-evaluation}), revealing a tension between abstractive generation and factual faithfulness, while demonstrating that  LLMs can surpass human baselines in strict evidence-grounding.
    \item We perform a fine-grained meta-evaluation (Section~\ref{sec:meta-evaluation}) of faithfulness metrics, proving the inadequacy of traditional reference-based metrics for this task and validating the use of LLM-based judges to reliably evaluate structured, long-form generation.
\end{itemize}

\section{OARelatedWork Dataset}
To obtain the final dataset, we first created a corpus of scientific articles. To establish our corpus, we used two sources, the \emph{CORE}~\cite{CORE, Knoth2012CORETA} and \emph{Semantic Scholar}~\cite{s2orc,Kinney2023TheSS} corpora.

The developers of \emph{CORE} supplied us with papers that had been converted into XML/TEI format using GROBID. There are 5.4 million documents up to the year 2020, each containing parsed full-text, bibliography, and other metadata such as title and authors. In contrast to Semantic Scholar data, these documents do not have an assigned year, field of study, or DOI. We used Microsoft Academic Graph (MAG)~\cite{mag} to obtain these fields. Also, the bibliography was not linked. There was no known ID of the paper mentioned, so we had to perform bibliography linking (more in Section~\ref{sec:bib_linking}).

Regarding the \emph{Semantic Scholar} corpus, we used both the S2ORC~\cite{s2orc} dataset and the papers available from the Semantic Scholar API~\cite{Kinney2023TheSS}. In total, it consisted of 129.2M\footnote{This is the sum of 101.9 (API) and 27.3 (S2ORC). Thus, it is a number with duplicates.} documents up to 2023. The size is much larger than that of CORE because it also contains abstract-only papers. About 90\% of papers are abstract-only.

We merged CORE and both Semantic Scholar corpora together\footnote{Preferring CORE version when there is a collision.}, applied document structure and citations transformations (see Appendix~\ref{sec:content_hier} and Appendix~\ref{sec:cit_spans}), deduplication\footnote{We utilized searcher described in Appendix~\ref{sec:searcher-for-bibliography-linking}.}, and content filtering (Appendix~\ref{sec:content_cleaning}), and obtained an \emph{OAPapers} collection of size 124.9M papers.

To represent each document, we use the JSON format, which contains \uline{a tree representation of the content of a document}. There are multiple hierarchy levels for sections, subsections, paragraphs, and sentences. We believe that this tree representation might be beneficial for future use cases, as it allows you to easily select parts of a document or tag content parts at the input of a model.\footnote{We also created CLI software, to better navigate in our data. It is available here \url{https://github.com/KNOT-FIT-BUT/OAPapersViewer}. }

\subsection{Bibliography Linking}\label{sec:bib_linking}
We perform bibliography linking for two reasons. First, we used a data source containing documents without identified bibliography (CORE). The second reason is to expand existing links. For this task, we used (i) \emph{MAG}, (ii) the \emph{Semantic Scholar citation graph}, and (iii) our \emph{own searcher}. 

We consider \emph{Semantic Scholar} graph as our main source; however, during processing, we discovered that we can obtain additional links using also \emph{MAG}. Finally, we used our \emph{own searcher} implementation even though its contribution was marginal. The number of identified bibliographic entries increased only by 0.6\,\%, which is about 11.60 million new links. To evaluate our search, we decided to perform a manual evaluation of the results and found that the search was correct in 47 of 50 cases. See Appendix~\ref{sec:searcher-for-bibliography-linking} for more details.

\subsection{Related Work Dataset}\label{sec:dataset}

To create a suitable dataset for the generation of related work sections, we search in our corpus for documents containing related work-like section. This means that we are not looking just for sections with the \emph{``Related Work''} headline, but also, e.g., \emph{``Background''}, \emph{``Related Literature''}, and \emph{``Literature Review''}. The \emph{``Background''} is only considered valid when there is another section named \emph{``Introduction''} to avoid confusion. 

In this way, we were able to find more than half a million documents with related work-like sections. However, we decided to filter them further to increase the quality of our final set. Our pipeline filters out each related work that (i) contains fewer than three sentences, (ii) cites fewer than two documents, or (iii) contains \emph{citation group} without a linked document.

We define a \emph{citation group} as a sequence of subsequent citation spans (e.g., \emph{[1][4][6]}). The condition to groups is used because having full coverage of cited papers is unrealistic in an open-access setup. We believe that having at least one representative of each group is a good approximation, as it is often the case that citations in a group are cited for similar reasons.

We also make sure that every target document, for which the related work is generated, and all cited documents have an abstract.

In this way, we obtain 94~450 papers with a related work section. (Table~\ref{tab:dataset_comparison} reports the average number of source documents per related work section).

Of all of these unique cited papers, 82\% are abstract-only papers. As we want to provide a dataset that is suitable for evaluation of models on full-texts, we decided to make sure that \emph{every related work section in validation and test set splits has all source documents with full-text}. To achieve that, we found all documents satisfying this condition and used 50\% of them as our test set, 30\% of them as our validation set, and we assigned the rest to the train set.

\begin{table}[htbp]
    \scriptsize 
    \centering
    \setlength{\tabcolsep}{3pt} 
    \begin{tabular}{l c r r r c}
        \toprule
        \textbf{Dataset} & \makecell{\textbf{Samples} \\ \textbf{(Tr/Va/Te)}} & \makecell{\textbf{Tgt.} \\ \textbf{Words}} & \makecell{\textbf{Cited} \\ \textbf{Words}} & \textbf{Docs} & \textbf{Para.} \\
        \midrule
        \makecell[l]{S2ORC \\ \cite{Chen2021CapturingRB}} & 126k / 5k / 5k & 148.0 & 1.1k & 5.0 & 1.0 \\
        \makecell[l]{Delve \\ \cite{Chen2021CapturingRB}} & 73k / 3k / 3k & 181.0 & 0.6k & 3.7 & 1.0 \\
        \makecell[l]{Multi-XSci. \\ \cite{Lu2020MultiXScienceAL}} & 30k / 5k / 5k & 116.4 & 0.8k & 4.4 & 1.0 \\
        \makecell[l]{STRoGeNS \\ \cite{structured_related_work}} & 86k / - / - & 514.3 & 3.0k & 16.6 & 4.2 \\
        \midrule
        \multicolumn{6}{l}{\textbf{OARelatedWork (Ours)}} \\
        Train & 91k & 530.3 & 26.2k & 11.4 & 4.5 \\
        Validation & 1.1k & 338.9 & 36.5k & 6.5 & 3.2 \\
        Test & 1.9k & 336.5 & 36.6k & 6.5 & 3.2 \\
        \bottomrule
    \end{tabular}
    \caption{Comparison of our dataset with existing related-work generation datasets.}
    \label{tab:dataset_comparison}
\end{table}

However, as can be seen in Table~\ref{tab:dataset_comparison}, this way of splitting caused the train set to contain longer targets, as it is more probable to cover all citations in shorter text, because shorter related works have fewer citations on average. Nevertheless, this does not present a fundamental problem given our task definitions (see Section~\ref{sec:tasks_definition}).

\paragraph{Domain Shift}
Our dataset exhibits a heavy shift toward computer science compared to the original corpus distribution (Appendix~\ref{sec:app-domain-shift}), likely because Computer Science papers explicitly section ``Related Work,'' whereas medicine and psychology integrate it into the introduction.

\section{Task Definition}\label{sec:tasks_definition}

A system is provided with (1) a combination $X$ of content types, including abstracts and full texts of the target paper (excluding the related work section) and its cited papers, (2) a target length $L$, and (3) information on whether the output should follow a structured format with subsections or a plain unstructured form.

Using these inputs, the task is to generate related work $Y = (y_1, y_2, ..., y_L)$, where $y_i$ is the i-th token of the related work section with the length $L$. Target length and subsection information are provided because they are subjective decisions of an author rather than a feature of the input text.

To identify the contribution of using different inputs, we experiment with multiple input type combinations (shown later in Test set results Table~\ref{tab:results}).

\section{Evaluation}\label{sec:evaluation}
We use several metrics to evaluate the generated text. To enable comparison with previous work, we use the ROUGE-1-F1, ROUGE-2-F1, and ROUGE-L-F1\footnote{We use the summary-level variant.} metrics~\cite{rouge}, and we also use BERTScore (BS)~\cite{bert-score}.

We conducted human evaluation for selected configurations and use it for both evaluating models and meta-evaluating LLM judges (Section~\ref{sec:human-evaluation}). From tested models we selected the quantized Gemma~4~31B \cite{gemma4} without reasoning as our judge. This judge extracts statements and assigns them to one of four categories: \textit{True}, \textit{True but wrong citation} (hereafter referred to as \textit{Wrong Cit.}), \textit{Unverifiable}, or \textit{False}. Further details are provided in Section~\ref{sec:meta-evaluation}.

Before applying an evaluation metric\footnote{This does not hold for citation metrics and LLM judges.}, we convert all citation and reference spans\footnote{E.g., references to figures and tables.} to form \texttt{<cite>} and \texttt{<ref>} from their original form \texttt{<cite> ID <sep> TITLE <sep> FIRST\_AUTHOR <sep> ... </cite>} and similar for references.

\noindent\textbf{Citation Metric}.
We additionally compute $F_1$ between the sets of cited papers in the reference and the generation (full definition in Appendix~\ref{app:citation-metric}).

\section{Related Work}
Our work aligns with the survey by \citet{li-ouyang-2024-related}. We benchmark their recommended baselines alongside modern LLMs, preserve complex multi-paper citations (Appendix~\ref{sec:cit_spans}), and explore Retrieval-Augmented Generation (RAG) performance bounds via our greedy oracle experiments (defined in Section~\ref{ss:ext_bsl_oracl}). Highlighting the field's shift toward full-text synthesis, \citet{liu-etal-2025-select} recently adopted the preprint of our OARelatedWork dataset to benchmark their novel multi-agent generation framework. Other recent works also explore full-text inputs, but at small scale: \citet{martinboyle2024shallowsynthesisknowledgegptgenerated} use GPT-4 to generate whole sections on 10 papers, and \citet{li-ouyang-2025-explaining} apply feature-based prompting on 27 papers. OARelatedWork fills the resulting need for a large-scale full-text whole-section benchmark.

\subsection{Datasets}
The very first work~\cite{Hoang2010TowardsAR} introduced related work generation with a dataset of 20 papers containing summary references, full-texts of cited articles, and topic trees, targeting whole related work sections. However, because abstractive approaches struggle with long inputs and outputs, subsequent datasets restricted inputs to abstracts and outputs to single paragraphs~\cite{Lu2020MultiXScienceAL, Chen2021CapturingRB}; a comparison with our data is available in Table~\ref{tab:dataset_comparison}.

A parallel line of work targets \emph{citation text generation}, producing a sentence (or a few) citing a given reference from inputs such as the title and abstract of the cited paper~\cite{AbuRaed2020AutomaticRW} or the citation context together with the cited abstract~\cite{xing-etal-2020-automatic}. \citet{funkquist-etal-2023-citebench} released \emph{CiteBench} to unify these datasets. \citet{li-etal-2024-cited} move beyond abstracts by retrieving cited text spans from the full text. 

Operating at a larger scale, SurveyGen~\cite{bao-etal-2025-surveygen} targets the generation of complete survey papers, but, unlike our work, still lacks the full texts of cited references.

Months after the pre-print version of our dataset was publicly released\footnote{Our dataset was released in May 2024, while \citet{structured_related_work} was in August 2024.}, \citet{structured_related_work} presented the \emph{STRoGeNS} dataset of structured summaries that, like ours, targets whole related work sections rather than single paragraphs. However, it uses only abstracts, leaving our test set with more than 12$\times$ more input data (Table~\ref{tab:dataset_comparison}). Nevertheless, \emph{STRoGeNS} has more cited documents on average. We believe that generating an entire related work section requires going beyond abstracts, when possible, and also providing the rest of the target paper to enable contrasting paragraphs (such as this one).

\subsection{Evaluation}\label{sec:related_work_evaluation}
Related work generation is a summarization task~\cite{Hoang2010TowardsAR} and inherits its evaluation challenges, including low inter-annotator agreement, especially for non-experts~\cite{Fabbri2020SummEvalRS, gillick-liu-2010-non}. Lexical-overlap metrics such as ROUGE~\cite{rouge}, BLEU~\cite{BLEU}, and METEOR~\cite{lavie-agarwal-2007-meteor} are widely used but fail to match semantically equivalent texts with different surface forms. Model-based metrics like BERTScore~\cite{bert-score} and BARTScore~\cite{bart_score} mitigate this but are limited by short context windows (e.g.,~512 tokens for BERT), which is problematic for our long-form task. More recently, \citet{geval} proposed \emph{G-Eval}, which prompts a large language model (GPT-4) with the task description, intermediate instructions, and a token-probability-based scoring function. Another worth mentioning is Prometheus 2 ~\cite{kim-etal-2024-prometheus}, an open-weight LLM dedicated to evaluation.

\paragraph{Claim-level faithfulness evaluation.} 
A complementary stream decomposes generations into atomic claims and verifies each one, starting with FActScore~\cite{min-etal-2023-factscore} for biographies and extended to summarization by FENICE~\cite{scire-etal-2024-fenice}, which combines claim extraction with NLI alignment, and MiniCheck~\cite{tang-etal-2024-minicheck}, which trains efficient grounded fact-checkers. For long inputs, LongDocFACTScore~\cite{bishop-etal-2024-longdocfactscore} and FaStFact~\cite{wan-etal-2025-fastfact} address the scalability of claim-based pipelines, while \citet{zhong-litman-2025-discourse} show that discourse-aware decomposition further improves the detection of factual inconsistency. FABLES~\cite{kim2024fablesevaluatingfaithfulnesscontent} performs claim-level human annotation of LLM-generated book-length summaries and finds that automatic raters correlate poorly with humans on long-form faithfulness. We adopt this paradigm in the related-work generation setting: our LLM judge extracts statements and assigns them to four citation-aware categories (\textit{True}, \textit{Wrong Cit.}, \textit{Unverifiable}, \textit{False}), and we meta-evaluate it against fine-grained human annotations (Section~\ref{sec:meta-evaluation}).

\section{Experiments}
We investigate several baselines on our dataset and ablate the impact of different types of information-rich inputs. These cover naive and oracle approaches, traditional systems, and transformer-based models.

\begin{table*}[ht!]
\scriptsize
\centering
\begin{tblr}{cccccccccccccc} 
    \hline
    \hline
    \SetCell[r=3]{c} \textbf{Model} & \SetCell[c=4]{c} \textbf{Input} & & & & \SetCell[r=3]{c} \textbf{citations} & \SetCell[r=3]{c} \textbf{R1} & \SetCell[r=3]{c} \textbf{R2} & \SetCell[r=3]{c} \textbf{RL} & \SetCell[r=3]{c} \textbf{BS} & \SetCell[r=3]{c} \textbf{True} & \SetCell[r=3]{c} \textbf{Wrong Cit.} & \SetCell[r=3]{c} \textbf{Unverifiable} & \SetCell[r=3]{c} \textbf{False}  \\
    \cline{2-5}
    & \SetCell[c=2]{c} \textbf{References} & & \SetCell[c=2]{c} \textbf{Target Paper} \\
    \cline{2-5}
    & \textbf{Abs.} & \textbf{Cont.} & \textbf{Abs.} & \textbf{Cont.} & & & & & & & & & \\
    \hline
LEAD & \checkmark &  &  &  & \textbf{99} & 26 & 4 & 20 & 5 & $99.1_{98.8}^{99.3}$ & $0.2_{0.1}^{0.4}$ & $0.6_{0.5}^{0.9}$ & $0.0_{0.0}^{0.0}$ \\
\hline
Greedy Oracle & \checkmark &  &  &  & 0 & 38 & 12 & 34 & 13 & $98.1_{97.6}^{98.5}$ & $0.1_{0.0}^{0.2}$ & $1.1_{0.8}^{1.4}$ & $0.6_{0.4}^{0.9}$ \\
 & \checkmark & \checkmark & \checkmark & \checkmark & 15 & \textbf{53} & \textbf{26} & \textbf{49} & \textbf{25} & $96.3_{95.7}^{96.7}$ & $0.7_{0.5}^{0.9}$ & $2.3_{1.9}^{2.7}$ & $0.3_{0.2}^{0.5}$ \\
 & \checkmark & \checkmark & \checkmark &  & 7 & 49 & 23 & 46 & 21 & $96.0_{95.3}^{96.5}$ & $0.4_{0.3}^{0.6}$ & $2.4_{2.0}^{2.8}$ & $0.4_{0.3}^{0.5}$ \\
 & \checkmark & \checkmark &  &  & 8 & 47 & 22 & 44 & 20 & $94.9_{94.2}^{95.5}$ & $0.3_{0.2}^{0.4}$ & $3.2_{2.8}^{3.7}$ & $0.4_{0.3}^{0.6}$ \\
 & \checkmark &  & \checkmark & \checkmark & 16 & 47 & 21 & 43 & 21 & $97.6_{97.1}^{98.0}$ & $0.4_{0.3}^{0.7}$ & $1.2_{1.0}^{1.5}$ & $0.5_{0.3}^{0.7}$ \\
\hline
TextRank & \checkmark &  &  &  & \textbf{99} & 31 & 6 & 29 & 9 & $99.4_{99.1}^{99.6}$ & $0.2_{0.1}^{0.5}$ & $0.3_{0.2}^{0.5}$ & $0.0_{0.0}^{0.1}$ \\
 & \checkmark & \checkmark & \checkmark & \checkmark & 65 & 31 & 5 & 27 & 7 & $96.7_{96.1}^{97.1}$ & $1.3_{1.1}^{1.7}$ & $1.0_{0.7}^{1.4}$ & $0.3_{0.2}^{0.4}$ \\
 & \checkmark & \checkmark & \checkmark &  & 67 & 34 & 6 & 30 & 9 & $97.5_{97.0}^{97.8}$ & $1.0_{0.8}^{1.3}$ & $0.6_{0.4}^{0.8}$ & $0.3_{0.2}^{0.6}$ \\
 & \checkmark & \checkmark &  &  & 68 & 29 & 4 & 25 & 6 & $95.1_{94.4}^{95.8}$ & $1.8_{1.5}^{2.2}$ & $1.3_{1.0}^{1.7}$ & $0.2_{0.1}^{0.3}$ \\
 & \checkmark &  & \checkmark & \checkmark & 89 & 33 & 7 & 30 & 11 & $99.5_{99.3}^{99.6}$ & $0.1_{0.1}^{0.2}$ & $0.3_{0.2}^{0.4}$ & $0.0_{0.0}^{0.1}$ \\
\hline
PRIMERA & \checkmark &  & \checkmark &  & 96 & 35 & 8 & 32 & 13 & $69.2_{68.4}^{70.1}$ & $4.5_{4.2}^{4.8}$ & $14.4_{13.8}^{15.0}$ & $10.2_{9.8}^{10.6}$ \\
 & GO & GO & \checkmark & GO & 93 & 42 & 15 & 39 & 18 & $65.1_{64.4}^{65.8}$ & $7.1_{6.7}^{7.5}$ & $12.3_{11.8}^{12.7}$ & $13.1_{12.6}^{13.6}$ \\
 & GO & GO & \checkmark &  & 94 & 41 & 13 & 37 & 17 & $63.4_{62.7}^{64.2}$ & $6.0_{5.7}^{6.3}$ & $14.3_{13.8}^{14.8}$ & $13.6_{13.2}^{14.1}$ \\
 & \checkmark &  & \checkmark & GO & 93 & 40 & 12 & 37 & 17 & $69.1_{68.3}^{69.8}$ & $5.7_{5.3}^{6.0}$ & $12.4_{12.0}^{12.9}$ & $11.7_{11.2}^{12.1}$ \\
\hline
Llama 3.2 3B & \checkmark &  & \checkmark &  & 2 & 29 & 6 & 26 & 5 & $83.5_{82.7}^{84.3}$ & $4.8_{4.4}^{5.3}$ & $6.8_{6.3}^{7.3}$ & $2.8_{2.5}^{3.1}$ \\
 & \checkmark & \checkmark & \checkmark & \checkmark & 9 & 23 & 4 & 21 & -1 & $66.9_{65.6}^{68.1}$ & $3.7_{3.4}^{4.2}$ & $23.4_{22.2}^{24.6}$ & $3.9_{3.6}^{4.2}$ \\
 & GO & GO & \checkmark & GO & 14 & 29 & 7 & 27 & 4 & $71.0_{70.0}^{72.0}$ & $5.8_{5.3}^{6.2}$ & $14.2_{13.5}^{15.0}$ & $6.2_{5.8}^{6.7}$ \\
\hline
GPT-4o Mini & \checkmark &  & \checkmark &  & 88 & 33 & 5 & 30 & 12 & $92.9_{92.3}^{93.4}$ & $2.8_{2.4}^{3.2}$ & $2.7_{2.4}^{3.0}$ & $0.8_{0.6}^{0.9}$ \\
 & \checkmark & \checkmark & \checkmark & \checkmark & 73 & 32 & 5 & 29 & 11 & $83.8_{82.9}^{84.6}$ & $6.0_{5.5}^{6.5}$ & $6.0_{5.6}^{6.5}$ & $1.2_{1.0}^{1.4}$ \\
 & GO & GO & \checkmark & GO & 78 & 35 & 6 & 31 & 14 & $81.4_{80.5}^{82.2}$ & $8.6_{8.0}^{9.3}$ & $5.8_{5.3}^{6.2}$ & $2.0_{1.8}^{2.3}$ \\
\hline
Human (Gold) & - & - & - & - & - & - & - & - & - & $73.5_{72.4}^{74.6}$ & $1.6_{1.4}^{1.9}$ & $18.5_{17.6}^{19.5}$ & $2.3_{2.1}^{2.6}$ \\
\hline
\end{tblr}
\caption{Test set results on multiple input variants. Inputs with Greedy Oracle (GO) are texts pre-extracted using Greedy Oracle extractor. Factuality metrics (True, Wrong Cit., Unverifiable, False) include 95\% bootstrap confidence intervals formatted as $Mean_{Lower}^{Upper}$. Missing percentages reflect judge output errors.}
\label{tab:results}
\vspace{-1em}
\end{table*}

\subsection{Models}\label{sec:models}

We evaluate three groups of models: extractive baselines and oracles, a fine-tuned multi-document summarization model, and zero-shot LLM baselines. Full implementation details, hyperparameters, and prompts are provided in the Appendix \ref{sec:primera-fine-tuning}, \ref{sec:used-tools}, and \ref{app:prompts}.

\begin{description}[style=unboxed,leftmargin=0em,listparindent=\parindent]
    \setlength\parskip{0em}

\item[Extractive Baselines and Oracles]\label{ss:ext_bsl_oracl}

We include standard extractive methods to establish baseline performance and estimate the upper bound of purely extractive summarization. \textsc{Lead} selects the first sentence from each input abstract and appends the corresponding citation span. \textsc{TextRank}~\cite{mihalcea-tarau-2004-textrank} ranks sentences using lexical-overlap-based PageRank and is applied separately to each input document; the resulting sub-summaries are concatenated and assigned citation spans.

We also use a \textsc{Greedy Oracle} (GO), which iteratively selects sentences that maximize ROUGE-2 F1 against the reference summary~\cite{Nallapati2016SummaRuNNerAR}. We evaluate GO both as a standalone extractive summarizer, where sentences are selected globally from all input documents, and as a retrieval component for neural models, where it produces one extractive sub-summary per document before concatenation.

\item[PRIMERA]

PRIMERA~\cite{primera} is an LED-based~\cite{Longformer} multi-document summarization model with 447M parameters, pretrained using Gap Sentence Generation and Entity Pyramid Masking. We fine-tune PRIMERA on our task and report the average over three runs. Input length statistics and fine-tuning details are provided in Appendices~\ref{sec:input-length-analysis} and~\ref{sec:primera-fine-tuning}.

\item[Zero-shot Models]

For zero-shot baselines, we evaluate GPT-4o mini~\cite{Hurst2024GPT4oSC} and Llama~3.2~3B~\cite{grattafiori2024llama3herdmodels}\footnote{\url{https://huggingface.co/meta-llama/Llama-3.2-3B-Instruct}}. As with PRIMERA, documents are represented using a Markdown-like format. The details are deffered to Appendix~\ref{app:prompts}.

\end{description}
\subsection{Results}\label{sec:results}
We overview the impact of different input types across models in Table~\ref{tab:results}.

\begin{description}[style=unboxed,leftmargin=0em,listparindent=\parindent]
    \setlength\parskip{0em}

\item[Citation Coverage of GO:] We omit citation concatenation for Greedy Oracle (GO) outputs to isolate how oracle sentences naturally cover cited documents. Using the full target yields higher citation scores, indicating that \uline{cited documents are frequently referenced in similar contexts outside the related work} section (e.g., introductions). Notably, 16\,\% of citations are covered by salient sentences extracted solely from the target papers (assuming abstracts contain no citations).

\item[Citation Formatting in Abstractive Models:] Advanced LLMs (e.g., GPT-4o Mini) and trained \uline{models exhibit high citation scores}, successfully synthesizing information across multiple papers rather than collapsing to a single source. Conversely, we observed that smaller models like Llama~3.2 often struggle to strictly adhere to the requested citation format despite explicit prompt instructions.

\item[Faithfulness \& The Extractive Illusion:] Our meta-evaluated LLM judge suggests a tension between extractive evidence preservation and abstractive generation. Extractive baselines obtain very high \emph{True} scores ($95$--$99\%$), likely because their outputs closely reuse source text; however, these scores should not be interpreted as evidence of better coherence or synthesis. By contrast, the \uline{human-written gold summaries receive a lower strict evidence-grounded \emph{True} score ($73.5\%$) and a higher \emph{Unverifiable} rate ($18.5\%$) than the extractive baselines. This does not imply that the human summaries are mostly false}: their \emph{False} rate remains low ($2.3\%$). In Appendix~\ref{app:human-unverifiable-statements-analyzes}, our manual analysis of the unverifiable human-written statements identifies three causes: missing citations, automated parsing errors, and claims not grounded in the available source material. Among the 23 unverifiable statements analyzed, 7 were due to missing citations, 6 to parsing errors, and 10 to ungrounded claims. Among abstractive systems, \uline{GPT-4o Mini achieves the highest evidence-grounded factuality}, while PRIMERA produces more than $10\%$ \emph{False} statements in several settings.

\item[Influence of Available Content \& Citation Conflation:] \uline{Expanding input context paradoxically degrades factual accuracy for abstractive LLMs}. Providing full texts instead of just abstracts caused GPT-4o Mini's \emph{True} score to drop from 92.9\% to 83.8\%, accompanied by spikes in \emph{Wrong Citation} (correct claims misattributed to the wrong source) and \emph{Unverifiable} rates. Applying GO as a dense Oracle RAG step mitigated noise for the smaller Llama~3.2 (improving factuality) but further degraded performance for GPT-4o Mini.

\end{description}

\section{Human Evaluation}\label{sec:human-evaluation}

We conducted a human evaluation of PRIMERA, GPT-4o-mini, and human-written reference sections across 40 randomly selected papers, yielding 120 related work sections.\footnote{Annotator recruitment and compensation are detailed in Appendix~\ref{sec:annotators}.} Both models generated text using the greedy oracle setup with all available references and target papers.

Evaluating generated related work sections is a high-cognitive-load task, requiring annotators to read and synthesize context from multiple source papers. Given the significant human labor costs involved, scaling the sample size beyond 40 target papers (120 sections) was practically constrained. To account for this and rigorously support our claims, we prominently report confidence intervals for evaluation metrics, demonstrating the statistical reliability of our findings.

To evaluate factuality, we extracted factual statements from each section using GPT-5 (Prompt~\ref{prompt:statement-extraction-for-human-evaluation}, Appendix~\ref{app:prompts}), which extracted verbatim spans and initially classified them as \textit{True} or \textit{False}. We mapped these statements back to the source text using fuzzy matching (Dice--S{\o}rensen coefficient, threshold 0.8), successfully recovering 84\% of statements (42\% as exact matches).

To manage annotation workload, we randomly sampled up to three statements per section. To improve discriminability, we additionally constructed an adversarial subset using GPT-5-mini and GPT-OSS-20B. Statements with conflicting factuality labels across the extraction models were flagged as adversarial candidates. We sampled up to three adversarial statements per section and added them to the pool, yielding 408 statements for evaluation. Notably, human-written statements comprised only 16.4\% of this adversarial subset.

We employed a pairwise evaluation format (Figure~\ref{fig:annotation-interface}, Appendix~\ref{sec:app-annotation-instructions}). First, at the \emph{statement level}, annotators classified statements into four categories: \textit{True} (supported by sources), \textit{True but wrong citation} (factually correct but attributed to the wrong source), \textit{False} (contradicted by sources), or \textit{Unverifiable} (ambiguous, nonsensical, or lacking evidence).

Next, at the \emph{document level}, annotators indicated their preference (or a tie) between the two sections across four dimensions: (i)~\textbf{Relevance}: which section more directly and thoroughly relates to the target paper's topic, contributions, and context; (ii)~\textbf{Faithfulness}: which section is more factually accurate and avoids incorrect claims; 
(iii)~\textbf{Language}: which section is clearer, more grammatically correct, and easier to read; and 
(iv)~\textbf{Overall Ranking}: the section preferred overall.

Detailed inter-annotator agreement scores are provided in Appendix~\ref{app:iaa}. Notably, the seemingly low agreement for \textit{Language} ($\kappa = 0.1707$) stems from the \textit{kappa paradox}: annotators overwhelmingly selected ``no preference,'' making strict disagreements exceedingly rare. Furthermore, agreement dropped notably on the adversarial subset ($\kappa = 0.4368$) compared to randomly selected statements ($\kappa = 0.6556$), confirming these edge cases are difficult for humans to verify. 

We observe \uline{higher agreement for fine-grained statement-level faithfulness ($\kappa = 0.6169$) than for coarse document-level Faithfulness ($\kappa = 0.4301$)}. This aligns with LongEval \cite{longeval}, which found that evaluating faithfulness at a finer granularity yields more reliable judgments for long texts, corroborating our decision to fact-check localized statements.

Table~\ref{tab:pairwise_human_eval} presents the pairwise results for document-level annotators. GPT-4o-mini overwhelmingly dominates, significantly outperforming both PRIMERA and human authors. This aligns with recent findings on LLM superiority in summarization \citep{pu2023summarizationalmostdead}. Notably, GPT-4o-mini's dominance holds across all attributes except \textit{Faithfulness}, where it statistically tied with PRIMERA. In Appendix~\ref{sec:statement_level_evaluation}, we further explore \textit{Faithfulness} at the statement level.

To understand these preferences, we analyzed the linguistic properties of the generated texts. Flesch Reading Ease scores indicate that \uline{GPT-4o-mini generates exceptionally dense, complex text} ($-1.05$), compared to humans ($23.58$) and PRIMERA ($32.27$). This rich vocabulary likely triggers a known annotator bias where judges inherently favor articulate, authoritative-sounding text \citep{gudibande2023false}. Additionally, we assessed extractiveness via ROUGE-2 overlap with the source texts. \uline{PRIMERA exhibits a strong extractive tendency} (0.9157), whereas \uline{human writers (0.6502) and GPT-4o-mini (0.5217) are significantly more abstractive}.

Our coarse evaluation reveals that PRIMERA is statistically indistinguishable from human authors across all four attributes. We hypothesize this parity is heavily driven by the Greedy Oracle framework providing an idealized, highly relevant context. While further ablation is required, this suggests that pairing a weaker sequence-to-sequence model with a strong retrieval mechanism may compensate for its generative limitations. Nevertheless, Appendix~\ref{sec:statement_level_evaluation} shows that PRIMERA falls behind under statement-level faithfulness evaluation.

\begin{table}[t]
    \centering
    \scriptsize
    \setlength{\tabcolsep}{4pt}
    \begin{tabular}{llcc}
        \toprule
        \textbf{Attribute} & \textbf{Model Pair (A vs. B)} & \textbf{TWR (A)} & \textbf{95\% CI} \\
        \midrule
        \multirow{3}{*}{\textbf{Preference}} 
        & \textbf{GPT-4o-mini} vs. PRIMERA & \textbf{0.825} & [0.725, 0.913] \\
        & \textbf{Human vs. PRIMERA}       & \textbf{0.456} & \textbf{[0.338, 0.581]} \\
        & \textbf{GPT-4o-mini} vs. Human   & \textbf{0.800} & [0.700, 0.894] \\
        \midrule
        \multirow{3}{*}{\textbf{Relevance}} 
        & \textbf{GPT-4o-mini} vs. PRIMERA & \textbf{0.756} & [0.656, 0.844] \\
        & \textbf{Human vs. PRIMERA}       & \textbf{0.488} & \textbf{[0.381, 0.600]} \\
        & \textbf{GPT-4o-mini} vs. Human   & \textbf{0.788} & [0.694, 0.869] \\
        \midrule
        \multirow{3}{*}{\textbf{Faithfulness}} 
        & \textbf{GPT-4o-mini vs. PRIMERA} & \textbf{0.600} & \textbf{[0.481, 0.713]} \\
        & \textbf{Human vs. PRIMERA}       & \textbf{0.475} & \textbf{[0.356, 0.588]} \\
        & \textbf{GPT-4o-mini} vs. Human   & \textbf{0.700} & [0.600, 0.800] \\
        \midrule
        \multirow{3}{*}{\textbf{Language}} 
        & \textbf{GPT-4o-mini} vs. PRIMERA & \textbf{0.631} & [0.581, 0.681] \\
        & \textbf{Human vs. PRIMERA}       & \textbf{0.506} & \textbf{[0.438, 0.575]} \\
        & \textbf{GPT-4o-mini} vs. Human   & \textbf{0.625} & [0.575, 0.675] \\
        \bottomrule
    \end{tabular}
    \caption{Pairwise human document level evaluation results based on True Win Rate (TWR; ties award 0.5 points). Winning models are highlighted in bold. When a tie occurs (the neutral baseline of 0.500 falls within the 95\% Bootstrap Confidence Interval), the entire row is bolded.}
    \label{tab:pairwise_human_eval}

\vspace{-2em}
\end{table}

To establish a reliable foundation for our LLM judge meta-evaluation, we performed a fine-grained, statement-level faithfulness evaluation. Because scanning large text volumes introduces high cognitive load, our evaluation process included a targeted meta-annotation round to resolve initial disagreements and correctly categorize unverifiable claims. The detailed methodology, analyzes of unverifiable statements, and full evaluation results (including the complete performance table) are deferred to Appendix~\ref{sec:statement_level_evaluation}. Ultimately, this fine-grained analysis demonstrates that \uline{GPT-4o-mini beats human-written related work in factuality}, a finding that aligns with \citet{pu2023summarizationalmostdead}, who similarly observed high human hallucination rates.

\subsection{Meta-Evaluation}\label{sec:meta-evaluation}

\begin{figure}
    \centering
    \begin{subfigure}[b]{1.0\linewidth}
        \centering
        \includegraphics[width=\textwidth]{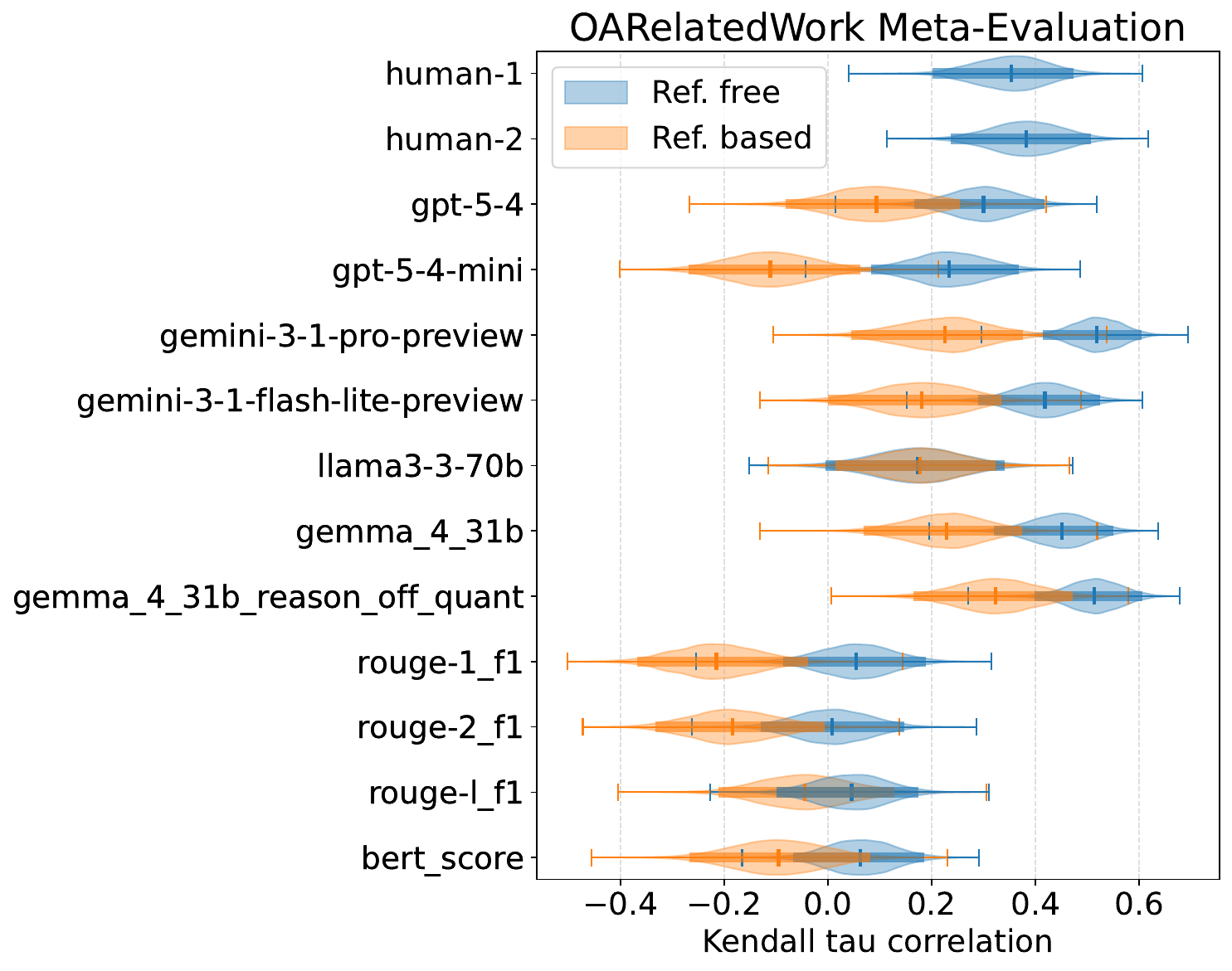}
    \end{subfigure}
    \caption{Violin plot of bootstrap distributions of Kendall's $\tau$ correlations of automatic metrics with human judgment on OARelatedWork meta-evaluation dataset for statement-level Faithfulness. Thick line is showing 95\% confidence interval.}
    \label{fig:coarse-violin-plot-correlation}
\vspace{-1em}
\end{figure}

We used the statement-level annotations to meta-evaluate candidate faithfulness metrics. We instructed LLM judges (Appendix~\ref{app:appendix_evaluator_models}, Prompt~\ref{prompt:the-judge-prompt}) to extract and classify statements, and to provide evidence and rationale, so that each decision becomes interpretable. We calculate the final coarse faithfulness as proportion of true statements in all extracted statements in given related work. This coarse-level transformation also allowed us to meta-evaluate LLM metrics together with metrics like ROUGE \cite{rouge} and BERTScore \cite{bert-score}. See Figure~\ref{fig:coarse-violin-plot-correlation} for the results.

Compared to LongEval baselines \cite{longeval}, non-LLM metrics performed poorly. Notably, the reference-based setup exhibited distribution shifts toward lower correlations, indicating that while they might suffice for validating abstract generation (see Appendix~\ref{app:comparison-to-longeval} for more details), \uline{the reference-based metrics are inadequate for related work generation}. 

Interestingly, the Figure~\ref{fig:coarse-violin-plot-correlation} shows that disabling reasoning for Gemma 4 did not degrade performance. We hypothesize this is because our output format inherently forces the model to provide evidence and a rationale, though confirming this requires further ablation. For the final evaluation, we utilized Gemma 4 without reasoning as a cost-effective alternative to Gemini 3.1 Pro.

We additionally report a meta-evaluation of F1 scores for pre-extracted statements in Appendix~\ref{app:meta-evaluation-detailed-results}.

\subsubsection{Parametric Knowledge Experiment}
To investigate reliance on internal parametric knowledge versus the strictness of context-bound evaluation, we conducted an ablation study restricting the judge model (Gemini 3.1 Pro) to verifying individual facts using only the greedy oracle inputs. Under these constraints, \textit{Unverifiable} statements increased across all sources. For GPT-4o-mini, they rose from 6.80\% to 17.69\%, indicating that even with an idealized context, the LLM moderately relies on its pre-trained weights to hallucinate external facts. Due to its extractive nature, PRIMERA exhibited the smallest shift (4.29\% to 7.86\%).

Most revealing was the human baseline: when evaluated strictly against the narrow greedy oracle extracts, its \textit{Unverifiable} rate increased from 19\,\% to 45\,\%. We conjecture, based on manual inspection, that \uline{human authors frequently integrate information distributed across the document} rather than relying on localized evidence spans. As a result, \uline{evaluation against strictly extractive context windows may systematically overestimate unverifiability} for human-written summaries.

We hypothesize that this ``Unverifiable delta'' could be used to empirically calibrate optimal chunk sizes and refine retrieval strategies of RAG pipelines.

\section{Conclusion and Future Work}
Our work shifts related-work generation toward a realistic, full-section paradigm by leveraging comprehensive full-text contexts. Our human evaluation provides insights into academic writing, revealing that human authors are highly abstractive and frequently introduce claims ungrounded in localized source texts. Consequently, advanced LLMs can actually outperform human writers in strict factual faithfulness. Furthermore, our meta-evaluation clearly demonstrates that standard reference-based metrics are insufficient for evaluating complex, long-form related work generation. By providing structured data, fine-grained factual annotations, and a specialized statement-level evaluation framework, we establish a robust foundation for future research.

\section*{Limitations and Ethical Considerations}
\paragraph{Data scope.} The dataset is limited to open-access papers and exhibits a heavy shift toward computer science (Section~\ref{sec:dataset}, Appendix~\ref{sec:app-domain-shift}).  Our findings about generator and judge behavior are therefore best calibrated to CS-style related work and may not transfer cleanly to other disciplines. The dataset is text-only, which might be limiting when the essential information is, e.g., in a plot.

\paragraph{Document processing.} Automatic parsing of PDFs into structured text is imperfect, and processing errors may miss or alter information. Our manual analysis (Appendix~\ref{app:human-unverifiable-statements-analyzes}) finds parsing-related errors in only \textasciitilde 5\% of human-written statements, indicating a reliable but not flawless pipeline.

\paragraph{Potential data contamination.} The papers in OARelatedWork are drawn from open-access corpora. It is therefore plausible that used LLM models have already seen used content.

\paragraph{Ethical use.} The baseline models are intended as research 
tools for professionals; their outputs should be reviewed critically.

\section*{Acknowledgement}
This work was supported by the EU CHIST-ERA project SoFAIR -- Making Software FAIR: Identifying and Extracting Software Mentions from Open Research Papers and Registering them with PIDs (the Technology Agency of the Czech Republic, Grant No. TH86010002, 2024-2025).

This work was supported by the Ministry of Education, Youth and Sports of the Czech Republic through the e-INFRA CZ (ID:90254)

\bibliography{custom}

\appendix
\section{Data Processing}\label{sec:data-processing}
\subsection{Searcher for Bibliography Linking}\label{sec:searcher-for-bibliography-linking}

Our search is done in multiple stages. The first stage is candidate retrieval. It consists of a string-match search on document titles using a SQLite database. We allow different additional postfixes for the matched title. Additionally, approximate matching is performed. It is looking for the k-nearest neighbors in the vector space of embeddings created using the feature locality-sensitive hashing~\cite{weinberger2010feature}. We used this method because of its low computational complexity. 

The last step is the filtering stage. The filter works with the title, authors, and year of a record. The title and authors fields are mandatory. If the year field is known for both records that are supposed to be matched, there cannot be more than a two-year difference. Following~\citet{s2orc}, we calculate the similarity score for a \emph{title} as the harmonic mean of the Jaccard index and containment metric:
\begin{equation}\label{containment_metric}
    C = \frac{|N_1 \cap N_2|}{\min(|N_1|, |N_2|)} \, ,
\end{equation}
where $N_1$ and $N_2$ are word sets from matched strings, which are also used for the Jaccard index.

Only titles with similarity above the $0.75$ threshold are considered as a match. When matching \emph{authors}, we require that at least one pair of authors have a match. The match is again done using a similarity metric (using the same threshold), but now it is just the containment metric because we want to obtain high similarity even when there is just, e.g., a surname in one source. We observe that it is often the case that the names are expressed using initials in one source and the full form in the other. For that reason, we also create an initial version of each name and consider it when matching.

For similarity computation via Jaccard / containment metric and string-match search, we first normalize the title and authors (authors are not used for string-match search) by converting the string into its close ASCII form using Unidecode\footnote{\url{https://pypi.org/project/Unidecode/}}, removing non-word characters\footnote{The \textbackslash W python regex placeholder.}, converting to lower case, and removing repeating characters.

\subsubsection{Estimation of the Error Effect of the Fallback Searcher}\label{sec:estimation-of-error-of-the-fallback-searcher}
Our manual 50-sample evaluation of the fallback searcher yielded an accuracy of 94\% (47/50 correct). As noted in Section~\ref{sec:bib_linking}, the fallback searcher contributed a marginal 0.6\% of the total identified links (the remaining 99.4\% were resolved via robust MAG and Semantic Scholar graphs). Under the conservative Clopper--Pearson exact method, the lower bound of the confidence interval for accuracy is 0.8345 (upper bound: 0.9875). Even under this worst-case statistical bound, the total potential error introduced to the overall dataset by the fallback searcher is strictly less than 0.1\% ($0.006 \times 0.1655 \approx 0.099\%$).

\subsection{Content Hierarchy}\label{sec:content_hier}

In the sources used, the documents were already parsed into sections and paragraphs. Nevertheless, we decided to add more levels to this shallow hierarchy, as we believe that it will increase the usability of our data. Ideally, we would like a document divided into sections, n-levels of subsections, paragraphs, and sentences. To segment sentences, we utilized the scispaCy model~\cite{neumann-etal-2019-scispacy}. 

The problem of parsing subsections is very difficult, as it is often the case that section headlines from existing resources are already not parsed correctly. We used a custom heuristic based on the numbering present in headlines. This heuristic falls back to a shallow hierarchy when parsing fails. More details are presented in Appendix~\ref{sec:subsection-parser}.

We manually analyzed our subsection parser on 100 papers. As we wanted only to evaluate the subsection parser, we only considered the provided headlines from the original corpora and not the true section headlines from the original PDF. We saw that for 20 papers, the algorithm failed and left the original shallow hierarchy. From the rest of the 80 papers, we considered 89\,\% parsed correctly. Our analysis also showed that 55\% of these 80 papers did not use subsections.

\subsection{Subsection Parser}\label{sec:subsection-parser}

\begin{figure}[h]
    \centering
    \includegraphics[width=8cm]{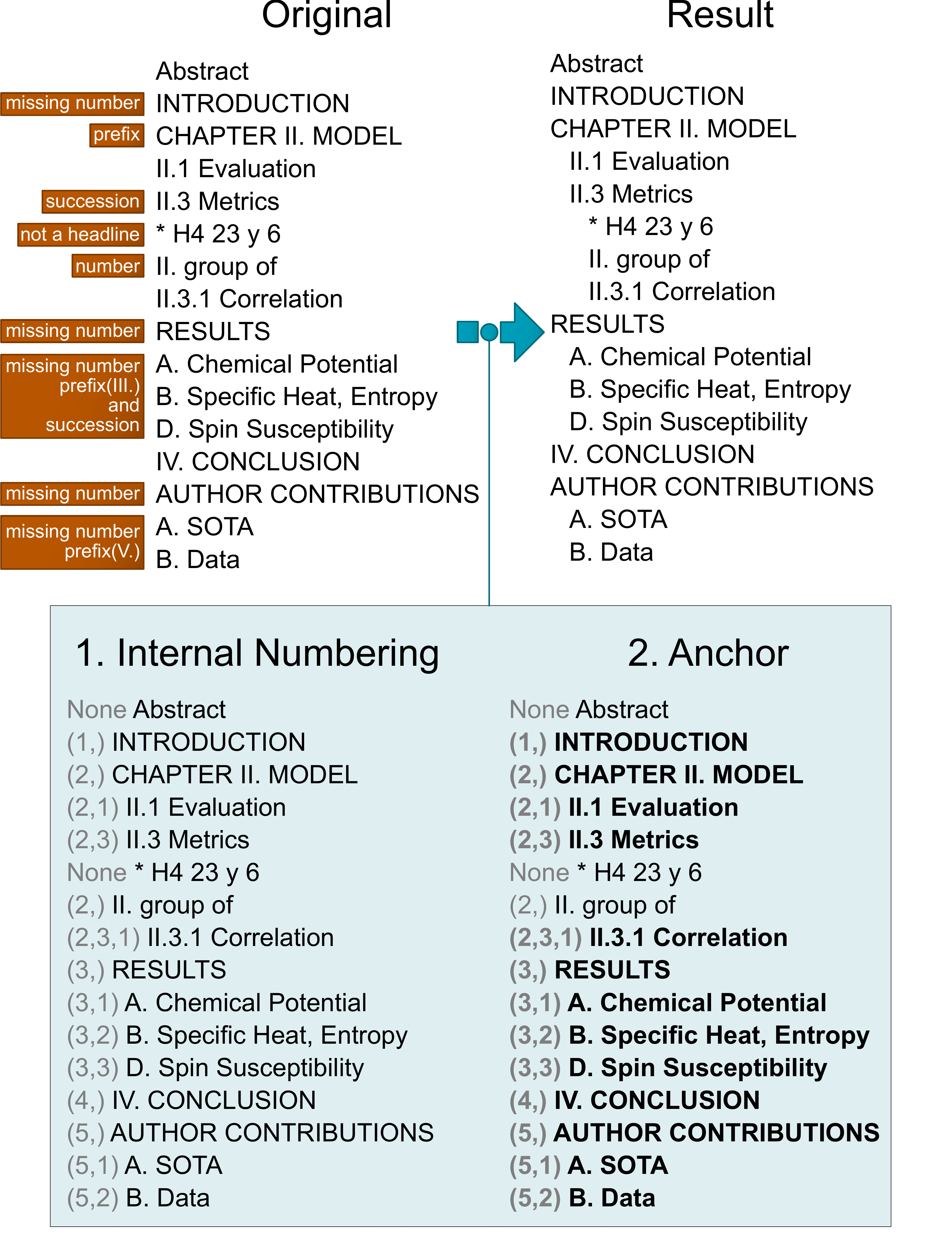}
    \caption{An example of hierarchy parsing showing two major steps: Internal Numbering and creation of Anchor. Orange boxes contain notes with obstacles for the parser. See that we are able to guess section numbers for headlines with missing numbers by exploiting writing habits and format. The introduction is usually the first section, and we can see that the capital letters format is used for all (known) top-level headlines. The headline \emph{* H4~23~y~6} will be removed in the cleaning step.} 
    \label{fig:hierarchy_parsing}
\end{figure}

To parse subsections, we use the numbering present in a headline. Our method is applied to the list of headlines, some of which might be subsections. It first tries to create an internal numbering by guessing section numbers for headlines with missing numbers by exploiting writing habits and format. See Figure~\ref{fig:hierarchy_parsing} for an example of conversion. Then comes identifying the longest sparse\footnote{We allow at most 3 missing numbers in between.} sequence of consecutive numberings (which we call the anchor). Then, it fits the rest of the headlines into a constructed hierarchy, as in the example in Figure~\ref{fig:hierarchy_parsing} where the non-bold headlines are put in the \emph{ II.3 metrics} section. The algorithm leaves the original, shallow hierarchy if it fails. The method stays with the original when the anchor consists of less than three elements or when the numbering is not matching levelwise. For example, assume that the previous subsection was 2.3.1 and that section 3.5 follows. It also considers the numeric format; thus e.g., for previous subsection 2.3.1, the following subsection 2.3.B would be considered problematic. Because an exhaustive description of the conversion would be beyond the scope of this paper, we direct the reader to investigate our code.

To perform sentence segmentation, we used the scispaCy model~\cite{neumann-etal-2019-scispacy}. It was trained on papers from the biomedical domain, so it is more suitable for our case as standard (web) models usually have problems with this type of document. One of the problems with standard models is that the citation spans are misclassified as sentence boundaries.

\subsection{Citation Spans}\label{sec:cit_spans}
In our data, we leave the citation spans in their original form, except for the merged citations (e.g., [3-6]). We decided to expand these to each individual citation, as we are providing character offsets to identify the citation span. Otherwise, it would mean that two citation spans could be in the same position.

The original sources already provide parsed citation spans. However, we decided to also employ our own simple parser for Harvard-style citations. It is a regular expression searcher that checks whether the information in searched spans corresponds to at least a single bibliographic entry. The matching is done on author and year. The year must match exactly, and the matching of the author is done in the same way as described in Section~\ref{sec:bib_linking}. This step is done after bibliography identification. We found an additional 13.66 million citation spans, which is a $1.8\%$ increase.

\subsection{Document Content Cleaning}\label{sec:content_cleaning}
As we are processing documents that might be badly parsed, we decided to add a cleaning step to our data processing pipeline. We remove:\footnote{By sections we also mean subsections.}
\begin{itemize}
    \setlength\itemsep{1pt}
    \item Sections with empty headline.
    \item Sections without text in their whole sub-hierarchy.
    \item Sections with headline with less than 30\% of Latin characters.
\end{itemize}

\section{Citation Metric}\label{app:citation-metric}

As citations are an important component of related work sections, we evaluate them separately. Let $R_c$ denote the set of cited documents in the reference (i.e., titles or identifiers of cited papers\footnote{We omit citations without resolvable bibliographic entries for clarity.}), and let $P_c$ denote the set of citations in the generated text. We measure citation quality using recall $c_r = |R_c \cap P_c| / |R_c|$ and precision $c_p = |R_c \cap P_c| / |P_c|$, and report their $F_1$ score.

\section{PRIMERA Fine-Tuning}\label{sec:primera-fine-tuning}

To fine-tune, we use the Lion~\cite{chen2024symbolic} optimizer with a linear scheduler with 4~000 warmup steps, learning rate 1e-05, no weight decay, batch size 16, and applied early stopping with patient 3. We use teacher forcing during training and beam search with beam size 4 during inference.

\begin{figure}[h]
    \centering
    \includegraphics[width=\linewidth]{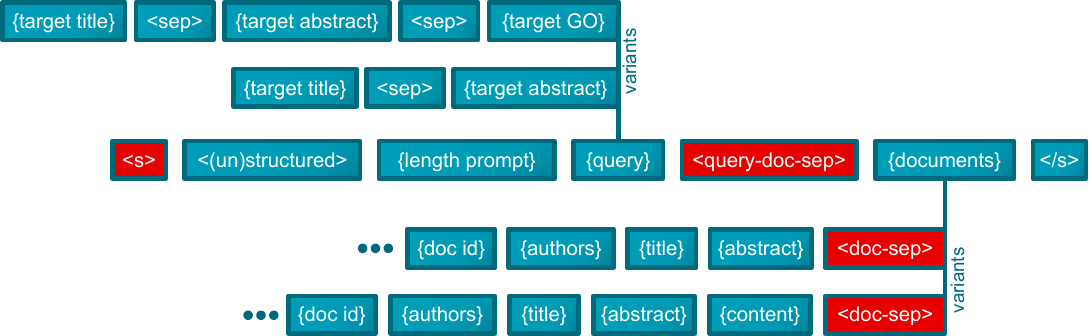}
    \caption{Illustration of input format for PRIMERA. The red boxes are global tokens. We put the special tokens \texttt{<unstructured>} or \texttt{<structured>} to identify whether we want to generate subsections. The \texttt{length prompt} is the number of tokens (in text form) of a reference target sequence. GO means content filtered by the Greedy Oracle model. In experiments, the GO is also used for filtering full-text of cited documents.}
    \label{fig:primera}
\end{figure}

Figure~\ref{fig:primera} shows the format of input. It can be seen that we added one new global token \texttt{<query-doc-sep>} for dividing query and content of cited documents. Document content, also target content, is represented using Markdown-like text with a special format of citations and references (see Section~\ref{sec:evaluation}). When content is not available, we put at least authors (max two) and title at the input. Each citation constitutes from id or unknown symbol, title of cited paper, and first author. This format allows models to draw the references from their memory.

Because we work with long inputs consisting of scientific papers (see Appendix~\ref{sec:input-length-analysis} for detailed length analysis) and the original model supports only inputs of maximal length 4~096 tokens and output of 1~024 tokens, we increased the number of encoder positional embeddings to 8~192 (16~384 for GO experiments) and 2~048 for the decoder. As in~\citet{Longformer} we initialized these additional embedings by copying the already existing ones.

\section{Input Length Analysis}\label{sec:input-length-analysis}
Figure~\ref{fig:lengths} shows the length analysis of different inputs.

\begin{figure*}[h]
    \centering
    \begin{subfigure}[t]{0.22\textwidth}
        \centering
        \caption{abstracts}
        \begin{tikzpicture}[font=\footnotesize]
    \begin{axis}[
        xbar,
        bar width=5pt,
        ylabel={tokens},
        xlabel={Counts},
        xlabel near ticks,
        ytick=data,
        ytick style={draw=none},
        yticklabel style = {xshift=5pt},
        yticklabels={
            136 - 2 213,
            2 214 - 4 291,
            4 292 - 6 369,
            6 370 - 8 447,
        },
        xmin=0,
        width=3.0cm,
        height=3.0cm,
        ticklabel style = {font=\tiny},
        ylabel near ticks,
        ylabel style = {font=\footnotesize},
        xlabel style = {font=\footnotesize},
        every x tick scale label/.style={at={(rel axis cs:1.0,0)},anchor=south west,inner sep=1pt},
        ]
        \addplot coordinates {
            (38 410, 1)
            (32 002, 2)
            (13 360, 3)
            (4 882, 4)
        };
    \end{axis}
\end{tikzpicture}
    \end{subfigure}
    ~
    \begin{subfigure}[t]{0.22\textwidth}
        \centering
        \caption{all}
        \begin{tikzpicture}[font=\footnotesize]
    \begin{axis}[
        xbar,
        bar width=5pt,
        xlabel={Counts},
        xlabel near ticks,
        ytick=data,
        ytick style={draw=none},
        yticklabel style = {xshift=5pt},
        yticklabels={
            670 - 32 956,
            32 957 - 65 243,
            65 244 - 97 530,
            97 531 - 129 817,
        },
        xmin=0,
        enlarge y limits=0.1,
        width=3.0cm,
        height=3.0cm,
        ticklabel style = {font=\tiny},
        ylabel near ticks,
        ylabel style = {font=\footnotesize},
        xlabel style = {font=\footnotesize},
        every x tick scale label/.style={at={(rel axis cs:1.1,0)},anchor=south west,inner sep=1pt},
        ]
        \addplot coordinates {
            (50 233, 1)
            (20 641, 2)
            (9 476, 3)
            (4 885, 4)
        };
    \end{axis}
\end{tikzpicture}
    \end{subfigure}
    ~
    \begin{subfigure}[t]{0.24\textwidth}
        \centering
        \caption{full cited + tar. abstract}
        \begin{tikzpicture}[font=\footnotesize]
    \begin{axis}[
        xbar,
        bar width=5pt,
        xlabel={Counts},
        xlabel near ticks,
        ytick=data,
        ytick style={draw=none},
        yticklabel style = {xshift=5pt},
        yticklabels={
            136 - 31 854,
            31 855 - 63 573,
            63 574 - 95 292,
            95 293 - 127 011,
        },
        xmin=0,
        enlarge y limits=0.1,
        width=3.0cm,
        height=3.0cm,
        ticklabel style = {font=\tiny},
        ylabel near ticks,
        ylabel style = {font=\footnotesize},
        xlabel style = {font=\footnotesize},
        every x tick scale label/.style={at={(rel axis cs:1.1,0)},anchor=south west,inner sep=1pt},
        ]
        \addplot coordinates {
            (56 959, 1)
            (16 672, 2)
            (7 941, 3)
            (4 263, 4)
        };
    \end{axis}
\end{tikzpicture}
    \end{subfigure}
    ~
    \begin{subfigure}[t]{0.21\textwidth}
        \centering
        \caption{full cited}
        \begin{tikzpicture}[font=\footnotesize]
    \begin{axis}[
        xbar,
        bar width=5pt,
        xlabel={Counts},
        xlabel near ticks,
        ytick=data,
        ytick style={draw=none},
        yticklabel style = {xshift=5pt},
        yticklabels={
            136 - 31 854,
            31 855 - 63 573,
            63 574 - 95 292,
            95 293 - 127 011,
        },
        xmin=0,
        enlarge y limits=0.1,
        width=3.0cm,
        height=3.0cm,
        ticklabel style = {font=\tiny},
        ylabel near ticks,
        ylabel style = {font=\footnotesize},
        xlabel style = {font=\footnotesize},
        every x tick scale label/.style={at={(rel axis cs:1.1,0)},anchor=south west,inner sep=1pt},
        ]
        \addplot coordinates {
            (56 959, 1)
            (16 672, 2)
            (7 941, 3)
            (4 263, 4)
        };
    \end{axis}
\end{tikzpicture}
    \end{subfigure}
    ~
    \begin{subfigure}[t]{0.22\textwidth}
        \centering
        \caption{abstracts + GO target}
        \begin{tikzpicture}[font=\footnotesize]
    \begin{axis}[
        xbar,
        bar width=5pt,
        ylabel={tokens},
        xlabel={Counts},
        xlabel near ticks,
        ytick=data,
        ytick style={draw=none},
        yticklabel style = {xshift=5pt},
        yticklabels={
            444 - 2 554,
            2 555 - 4 665,
            4 666 - 6 776,
            6 777 - 8 887,
        },
        xmin=0,
        enlarge y limits=0.1,
        width=3.0cm,
        height=3.0cm,
        ticklabel style = {font=\tiny},
        ylabel near ticks,
        ylabel style = {font=\footnotesize},
        xlabel style = {font=\footnotesize},
        every x tick scale label/.style={at={(rel axis cs:1.1,0)},anchor=south west,inner sep=1pt},
        ]
        \addplot coordinates {
            (29 638, 1)
            (33 684, 2)
            (16 657, 3)
            (6 992, 4)
        };
    \end{axis}
\end{tikzpicture}
    \end{subfigure}
    ~
    \begin{subfigure}[t]{0.22\textwidth}
        \centering
        \caption{abstracts + target}
        \begin{tikzpicture}[font=\footnotesize]
    \begin{axis}[
        xbar,
        bar width=5pt,
        xlabel={Counts},
        xlabel near ticks,
        ytick=data,
        ytick style={draw=none},
        yticklabel style = {xshift=5pt},
        yticklabels={
            670 - 11 783,
            11 784 - 22 897,
            22 898 - 34 011,
            34 012 - 45 125,
        },
        xmin=0,
        enlarge y limits=0.1,
        width=3.0cm,
        height=3.0cm,
        ticklabel style = {font=\tiny},
        ylabel near ticks,
        ylabel style = {font=\footnotesize},
        xlabel style = {font=\footnotesize},
        every x tick scale label/.style={at={(rel axis cs:1.1,0)},anchor=south west,inner sep=1pt},
        ]
        \addplot coordinates {
            (61 836,1)
            (25 803,2)
            (2 531,3)
            (620,4)
        };
    \end{axis}
\end{tikzpicture}
    \end{subfigure}
    ~
    \begin{subfigure}[t]{0.24\textwidth}
        \centering
        \caption{GO cited + tar. abstract}
        \begin{tikzpicture}[font=\footnotesize]
    \begin{axis}[
        xbar,
        bar width=5pt,
        xlabel={Counts},
        xlabel near ticks,
        ytick=data,
        ytick style={draw=none},
        yticklabel style = {xshift=5pt},
        yticklabels={
            122 - 9 265,
            9 266 - 18 409,
            18 410 - 27 553,
            27 554 - 36 697,
        },
        xmin=0,
        enlarge y limits=0.1,
        width=3.0cm,
        height=3.0cm,
        ticklabel style = {font=\tiny},
        ylabel near ticks,
        ylabel style = {font=\footnotesize},
        xlabel style = {font=\footnotesize},
        every x tick scale label/.style={at={(rel axis cs:1.1,0)},anchor=south west,inner sep=1pt},
        ]
        \addplot coordinates {
            (75 304,1)
            (11 351,2)
            (3 075,3)
            (1 040,4)
        };
    \end{axis}
\end{tikzpicture}
    \end{subfigure}
    ~
    \begin{subfigure}[t]{0.21\textwidth}
        \centering
        \caption{outputs}
        \begin{tikzpicture}[font=\footnotesize]
    \begin{axis}[
        xbar,
        bar width=5pt,
        ylabel={tokens},
        xlabel={Counts},
        xlabel near ticks,
        ytick=data,
        ytick style={draw=none},
        yticklabel style = {xshift=5pt},
        yticklabels={
            80 - 1 455,
            1 456 - 2 831,
            2 832 - 4 207,
            4 208 - 5 583,
        },
        xmin=0,
        enlarge y limits=0.1,
        width=3.0cm,
        height=3.0cm,
        ticklabel style = {font=\tiny},
        ylabel near ticks,
        ylabel style = {font=\footnotesize},
        xlabel style = {font=\footnotesize},
        every x tick scale label/.style={at={(rel axis cs:1.1,0)},anchor=south west,inner sep=1pt},
        ]
        \addplot coordinates {
            (75 314,1)
            (14 268,2)
            (1 393,3)
            (315,4)
        };
    \end{axis}
\end{tikzpicture}
    \end{subfigure}
    \caption{Histograms showing lengths of input types and lengths of outputs on train set. These numbers are obtained from PRIMERA tokenized inputs/outputs. GO for target paper is not created from abstract. GO summaries of cited documents are created for each document separately. We cut the top part of histograms to fit them on the page, thus the maximum is not there.}
    \label{fig:lengths}
\end{figure*}

\section{Other Used Tools}\label{sec:used-tools}
This section describes libraries, frameworks, or tools used during the creation of this work that were not already mentioned.

We use the Hugging Face \cite{Wolf_Transformers_State-of-the-Art_Natural_2020} library for obtaining ROUGE. We also used Hugging Face implementation of PRIMERA model. The TextRank implementation is taken from the summa Python package \cite{DBLP:journals/corr/BarriosLAW16}. We use Ollama \cite{ollama} for the Llama 3.2 3B. The GPT-4o Mini results were obtained using the official OpenAI API \url{https://openai.com/}.

\section{Domain shift}\label{sec:app-domain-shift}
This section provides the detailed distribution of study fields across the original corpus and our resulting dataset splits. As shown in Figure~\ref{fig:fos_distribution_shift}, while the original un-filtered corpus covers diverse disciplines (such as medicine and biology), the structural requirement of having dedicated related work-like sections introduces a pronounced bias. Consequently, computer science heavily dominates the final training, validation, and test sets.
\begin{figure}[h]
    \centering
    \begin{subfigure}[t]{0.8\columnwidth}
        \begin{tikzpicture}
\begin{axis}[
    xbar,
    xmin=0,
    xlabel={Original},
    xlabel style={
        at={(axis description cs:0.5,1.25)},
        anchor=south
    },
    every x tick scale label/.style={
            at={(xticklabel* cs:1.1,-0.1cm)},
            anchor=near xticklabel
    },
    symbolic y coords={engineering,chemistry,comp. science,biology,medicine},
    ytick=data,
    y tick label style={font=\footnotesize},
    legend style={at={(0.5,1.05)},
    anchor=south,legend columns=-1},
    enlarge y limits=0.15,
    bar width=10pt,
    width=5cm,
    height=4cm,
    tick pos=left,
    ticklabel style = {font=\tiny},
    ylabel style = {font=\footnotesize},
    xlabel style = {font=\footnotesize},
]

\addplot coordinates {
    (11327741,engineering)
    (13167138,chemistry)
    (14079372,comp. science)
    (18243662,biology)
    (36114766,medicine)
};

\end{axis}
\end{tikzpicture}
        \vspace{-25pt}
    \end{subfigure}
    \begin{subfigure}[t]{0.8\columnwidth}
        \begin{tikzpicture}
\begin{axis}[
    xbar,
    xmin=0,
    xlabel={Train},
    xlabel style={
        at={(axis description cs:0.5,1.25)},
        anchor=south
    },
    every x tick scale label/.style={
            at={(xticklabel* cs:1.1,-0.1cm)},
            anchor=near xticklabel
    },
    symbolic y coords={ml,nlp,mathematics,medicine,comp. science},
    ytick=data,
    y tick label style={font=\footnotesize},
    legend style={at={(0.5,-0.15)},
    anchor=north,legend columns=-1},
    enlarge y limits=0.15,
    width=5cm,
    height=4cm,
    bar width=10pt,
    tick pos=left,
    ticklabel style = {font=\tiny},
    ylabel style = {font=\footnotesize},
    xlabel style = {font=\footnotesize},
]

\addplot coordinates {
    (6367,ml)
    (6455,nlp)
    (9187,mathematics)
    (9564,medicine)
    (71069,comp. science)
};

\end{axis}
\end{tikzpicture}
        \vspace{-25pt}
    \end{subfigure}
    \begin{subfigure}[t]{0.8\columnwidth}
        \begin{tikzpicture}
\begin{axis}[
    xbar,
    xmin=0,
    xlabel={Validation / Test},
    xlabel style={
        at={(axis description cs:0.5,1.25)},
        anchor=south
    },
    scaled x ticks={base 10:-3},
    every x tick scale label/.style={
            at={(xticklabel* cs:1.10,-0.08cm)},
            anchor=near xticklabel
    },
    symbolic y coords={mach. trans.,mach. learn.,tpm,nlp,comp. science},
    ytick=data,
    y tick label style={font=\footnotesize},
    legend style={at={(0.62,0.30)},font=\tiny,
    anchor=north,legend columns=-1},
    enlarge y limits=0.15,
    width=5cm,
    height=4cm,
    bar width=3pt,
    tick pos=left,
    ticklabel style = {font=\tiny},
    ylabel style = {font=\footnotesize},
    xlabel style = {font=\footnotesize},
]

\addplot coordinates {
    (1016,comp. science)
    (333,nlp)
    (120,tpm)
    (0,mach. learn.)
    (112,mach. trans.)
};

\addplot coordinates {
    (1670,comp. science)
    (545,nlp)
    (211,tpm)
    (174,mach. learn.)
    (169,mach. trans.)
};

\legend{Validation,Test}

\end{axis}
\end{tikzpicture}
    \end{subfigure}
    \setlength{\abovecaptionskip}{-20pt}
    \caption{Field of study domain shift between all papers in used corpus and dataset splits. As a single paper may have multiple fields of study, the counts do not add up to the total number of papers. The \emph{tpm} is \emph{task (project management)}.}
    \label{fig:fos_distribution_shift}

\end{figure}

\section{Annotation Instructions}\label{sec:app-annotation-instructions}

\subsection{Objective}
The goal of this task is to \textbf{evaluate the quality of Related Work sections} for research papers.

\subsection{Materials Provided}
\begin{itemize}
    \item \textbf{Target Paper:} The main paper for which you are evaluating related work.
    \item \textbf{Two Versions of Related Work:} Competing drafts. You will select the version you consider better.
    \item \textbf{Cited Papers:} References to support citations in the related work sections.
\end{itemize}

\subsection{Annotation Process}

\paragraph{1. Familiarize Yourself}
Read the \textbf{abstract of the target paper} to understand its main contributions and scope.

\paragraph{2. Review Related Work Sections}
Carefully read \textbf{both versions} of the related work section.

\paragraph{3. Annotate Factual Statements}
\begin{itemize}
    \item There are \textbf{up to 6 factual statements} for each related work section.
    \item For each statement, choose:
    \begin{itemize}
        \item \textbf{True} -- The statement is supported by the target or cited papers.
        \item \textbf{True, but wrong citation} -- The statement is factually correct and supported by a different provided source, but not by the one it cites.
        \item \textbf{False} -- The statement contradicts the provided sources.
        \item \textbf{Unverifiable} -- The statement cannot be verified from the sources, is ambiguous, or nonsensical.
    \end{itemize}
    \item \textbf{Important:}
    \begin{itemize}
        \item Use information \textbf{only} from the target paper and cited papers.
        \item If a given statement span is not correct, create a new one and refer this new span to the original one (you can use \texttt{Alt+R}).
        \item Use the metadata field (\texttt{Alt+M}) if you want to add a comment about your decision.
        \item If you do not understand, select ``Unverifiable'' and write ``I do not understand'' in the metadata field (\texttt{Alt+M}).
    \end{itemize}
\end{itemize}

\paragraph{4. Select Preferred Related Work}
After reviewing all factual statements, decide which of the two provided related work sections you prefer for each evaluation aspect.

\vspace{0.5em}
\noindent \textbf{Evaluation Aspects:}
\begin{itemize}
    \item \textbf{Relevance} -- Choose the related work section that more directly and thoroughly relates to the target paper's topic, contributions, and context.
    \item \textbf{Faithfulness} -- Choose the section that is more factually accurate, avoids incorrect claims, and stays true to the source material.
    \item \textbf{Language} -- Choose the section that is clearer, more grammatically correct, and easier to read.
    \item \textbf{Overall Ranking} -- Choose \textbf{the related work section you prefer} overall out of the two given.
\end{itemize}

\subsection{Additional Notes}
\begin{itemize}
    \item Every \textbf{three consecutive tasks} share the same target paper; only the \textbf{two competing related works} change.
\end{itemize}

\section{Annotation Interface}\label{sec:app-annotation-interface}
\begin{figure*}
    \centering
    \includegraphics[width=0.9\linewidth]{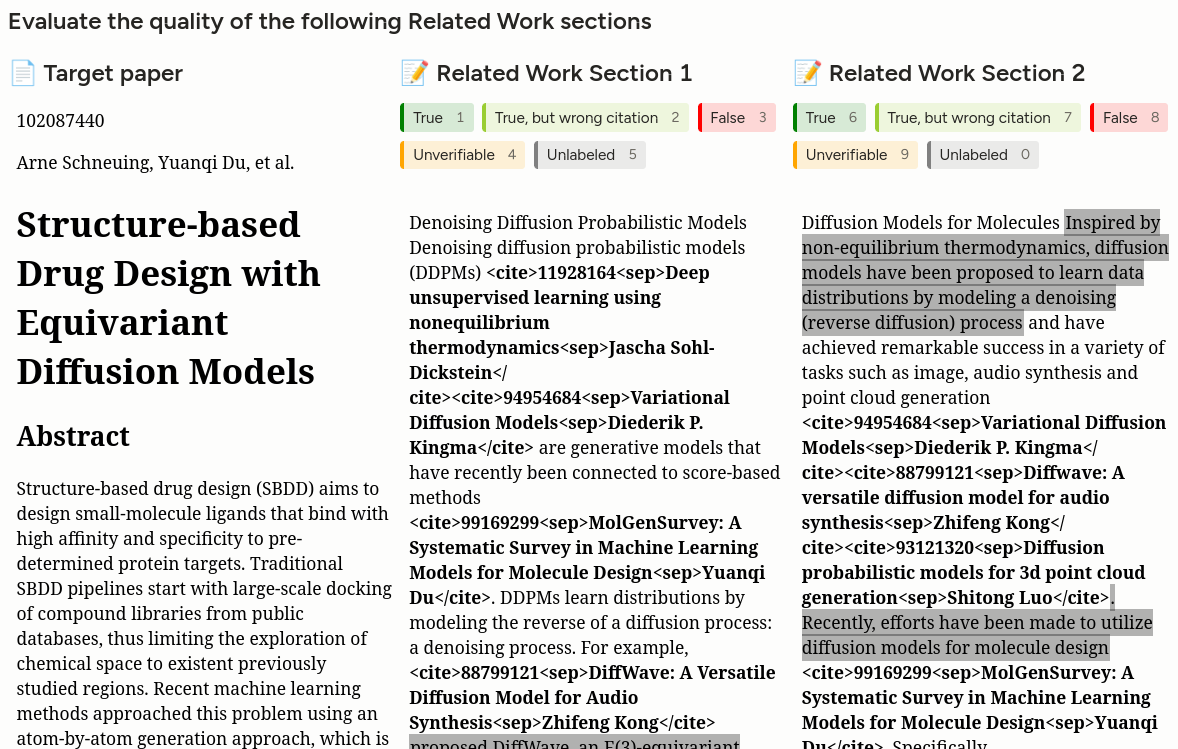}
    \caption{Annotation interface}
    \label{fig:annotation-interface}
\end{figure*}

The figure~\ref{fig:annotation-interface} shows the top portion of the annotation interface, specifically focusing on the statement annotation task. The layout presents the target paper on the left alongside two competing related work sections. Below these texts, the interface provides a dedicated area for the comparative evaluation aspects, followed by the full content of all cited papers for easy reference.

\subsection{Evaluator Models}
\label{app:appendix_evaluator_models}

To ensure a comprehensive and robust evaluation of automated fact-checking capabilities, we employed a diverse suite of Large Language Models (LLMs) to act as judges. Our selection spans state-of-the-art proprietary models accessed via official APIs and high-performing open-weight models deployed locally. For the proprietary suite, we evaluated the latest iterations of the GPT family (GPT-5.4 and GPT-5.4-mini) and the Gemini family (\texttt{gemini-3-1-pro-preview} and \texttt{gemini-3-1-flash-lite-preview}).

For the open-weight suite, we utilized the \texttt{vLLM} framework to optimize local inference throughput. We evaluated the 70-billion parameter Llama 3.3 (\texttt{meta-llama/Llama-3.3-70B-Instruct}) and the 31-billion parameter Gemma 4 (\texttt{google/gemma-4-31B-it}). Furthermore, to investigate the computational trade-offs of quantization and chain-of-thought generation, we included an AWQ-quantized variant of Gemma 4 (\texttt{QuantTrio/gemma-4-31B-it-AWQ}). Notably, because the Gemma 4 architecture natively generates explicit reasoning traces, we evaluated this quantized checkpoint with its reasoning generation explicitly disabled. This allowed us to assess whether removing the computational overhead of chain-of-thought reasoning, combined with quantization, degraded the model's strict fact-checking performance. The complete list of evaluator models and their corresponding endpoints or repositories is detailed in Table~\ref{tab:evaluator_models_details}.

\begin{table}[h]
    \centering
    \scriptsize
    \begin{tabular}{lll}
        \toprule
        \textbf{Family} & \textbf{Specific Checkpoint / Version} & \textbf{Access Type} \\
        \midrule
        \multirow{2}{*}{\textbf{GPT}} 
        & GPT-5.4 & Prop. API \\
        & GPT-5.4-mini & Prop. API \\
        \midrule
        \multirow{2}{*}{\textbf{Gemini}} 
        & \texttt{gemini-3-1-pro-preview} & Prop. API \\
        & \texttt{gemini-3-1-flash-lite-preview} & Prop. API \\
        \midrule
        \textbf{Llama} 
        & \texttt{meta-llama/Llama-3.3-70B-Instruct} & Open (vLLM) \\
        \midrule
        \multirow{2}{*}{\textbf{Gemma}} 
        & \texttt{google/gemma-4-31B-it} & Open (vLLM) \\
        & \texttt{QuantTrio/gemma-4-31B-it-AWQ} & Open (vLLM) \\
        \bottomrule
    \end{tabular}
    \caption{Details of the LLM evaluators used in the statement-level factuality assessment. Open-weight models were executed locally using the vLLM engine, while proprietary models were accessed via their respective official APIs.}
    \label{tab:evaluator_models_details}
\end{table}

\section{Inter-Annotator Agreement Details}
We provide inter-annotator agreements in Table~\ref{tab:kappa}.

\label{app:iaa}

\begin{table}[t]
    \centering
    \small
    \setlength{\tabcolsep}{6pt}
    \begin{tabular}{lc}
        \toprule
        \textbf{Aspect} & $\boldsymbol{\kappa}$ \\
        \midrule
        Preference & 0.5099 \\
        Relevance & 0.5452 \\
        Language & 0.1707 \\
        Faithfulness & 0.4301 \\
        Statement (all) & 0.6169 \\
        Statements (randomly selected) & 0.6556 \\
        Statements (randomly selected adversarial) & 0.4368 \\
        \bottomrule
    \end{tabular}
    \caption{Inter-annotator agreement. We use quadratic weighted Cohen's $\kappa$ for preference-based judgments (Preference/Relevance/Faithfulness/Language) because annotators could select \textit{no preference}. We use standard Cohen's $\kappa$ for statement labels (Statements).}
    \label{tab:kappa}
\end{table}

\section{Statement Level Evaluation}\label{sec:statement_level_evaluation}
We focus on fine-grained statement-level faithfulness, as it yields higher inter-annotator agreement (Table~\ref{tab:kappa}) and serves as the foundation for our LLM judge meta-evaluation.

Initial fine-grained annotations yielded high rates of \textit{Unverifiable} statements (41.3\% for humans, 26.5\% for GPT-4o-mini, 22.9\% for PRIMERA). To address this, we conducted a meta-annotation round for samples lacking consensus between human annotators and the original statement extraction model. A single annotator, different from the previous two, reviewed these cases with the original annotations and the model-provided evidence, significantly decreasing the \textit{Unverifiable} rate (Table~\ref{tab:fine_grained_factuality}). This suggests the initial round missed evidence due to the high cognitive load of scanning large text volumes. We provide detailed analyzes of unverifiable statements for human related works in Appendix~\ref{app:human-unverifiable-statements-analyzes}. Consequently, the initial high proportion of unverifiable claims likely influenced the coarse-level document ratings. The adversarial subset also consistently exhibited higher \textit{False} and \textit{Unverifiable} rates than the random subset, highlighting the difficulty of verifying these edge cases. 

Overall, \uline{GPT-4o-mini beats human-written related work in factuality}, aligning with \citet{pu2023summarizationalmostdead}, who similarly observed high human hallucination rates.

\begin{table}[t]
    \centering
    \small
    \setlength{\tabcolsep}{4pt}
    \begin{tabular}{lccccc}
        \toprule
        \textbf{Model} & \textbf{N} & \textbf{True} & \textbf{W. Cite} & \textbf{False} & \textbf{Unver.} \\
        \midrule
        \multicolumn{6}{c}{\textit{All Statements}} \\
        \midrule
        PRIMERA       & 140 & 58.6 & 5.7 & 30.7 & \textbf{5.0} \\
        Human         & 121 & 73.6 & \textbf{0.8} & 6.6  & 19.0 \\
        GPT-4o-mini   & 147 & \textbf{78.2} & 2.7 & \textbf{5.4}  & 13.6 \\
        \midrule
        \multicolumn{6}{c}{\textit{Randomly Selected}} \\
        \midrule
        PRIMERA       & 115 & 61.7 & 5.2 & 28.7 & \textbf{4.4} \\
        Human         & 110 & 76.4 & \textbf{0.9} & 3.6  & 19.1 \\
        GPT-4o-mini   & 116 & \textbf{82.8} & 2.6 & \textbf{2.6}  & 12.1 \\
        \midrule
        \multicolumn{6}{c}{\textit{Adversarial Subset}} \\
        \midrule
        PRIMERA       & 25  & 44.0 & 8.0 & 40.0 & \textbf{8.0} \\
        Human         & 11  & 45.5 & \textbf{0.0} & 36.4 & 18.2 \\
        GPT-4o-mini   & 31  & \textbf{61.3} & 3.2 & \textbf{16.1} & 19.4 \\
        \bottomrule
    \end{tabular}
    \caption{Fine-grained factuality evaluation. Metrics (aside from the total statement count, $N$) are reported as percentages (\%). "W. Cite" denotes statements that are true but have the wrong citation, and "Unver." denotes unverifiable statements. The best performing model (highest True and lowest False) is in bold.}
    \label{tab:fine_grained_factuality}
\end{table}

\section{Analysis of Unverifiable Statements in Human-Written Summaries}\label{app:human-unverifiable-statements-analyzes}

Because the proportion of unverifiable statements in human-authored summaries remained the highest even after the meta-annotation round, we conducted a deeper manual analysis of these statements. For this analysis, we reused the same subset of statements evaluated during the meta-annotation phase. 

We found that 23 out of 121 statements (19.0\%) in the human-written summaries were classified as unverifiable. Upon closer inspection of these 23 statements, we identified three primary causes for unverifiability: 

First, 7 statements (30.4\%) were unverifiable due to missing citations \footnote{This includes instances where the relevant source was omitted entirely, or cited elsewhere in the target paper outside of the related work section.}. Second, 6 statements (26.1\%) were flagged as unverifiable strictly due to automated parsing errors during extraction. Finally, 10 statements (43.5\%) contained information or claims that were fundamentally ungrounded in the source material.

This also reflects the quality of the processing pipeline: only 6 out of 121 statements were identified as problematic due to the pipeline itself, corresponding to a success rate of approximately 95\%.

\section{Comparison to LongEval}\label{app:comparison-to-longeval}
\begin{figure}
    \centering
    \begin{subfigure}[b]{1.0\linewidth}
        \centering
        \includegraphics[width=\textwidth]{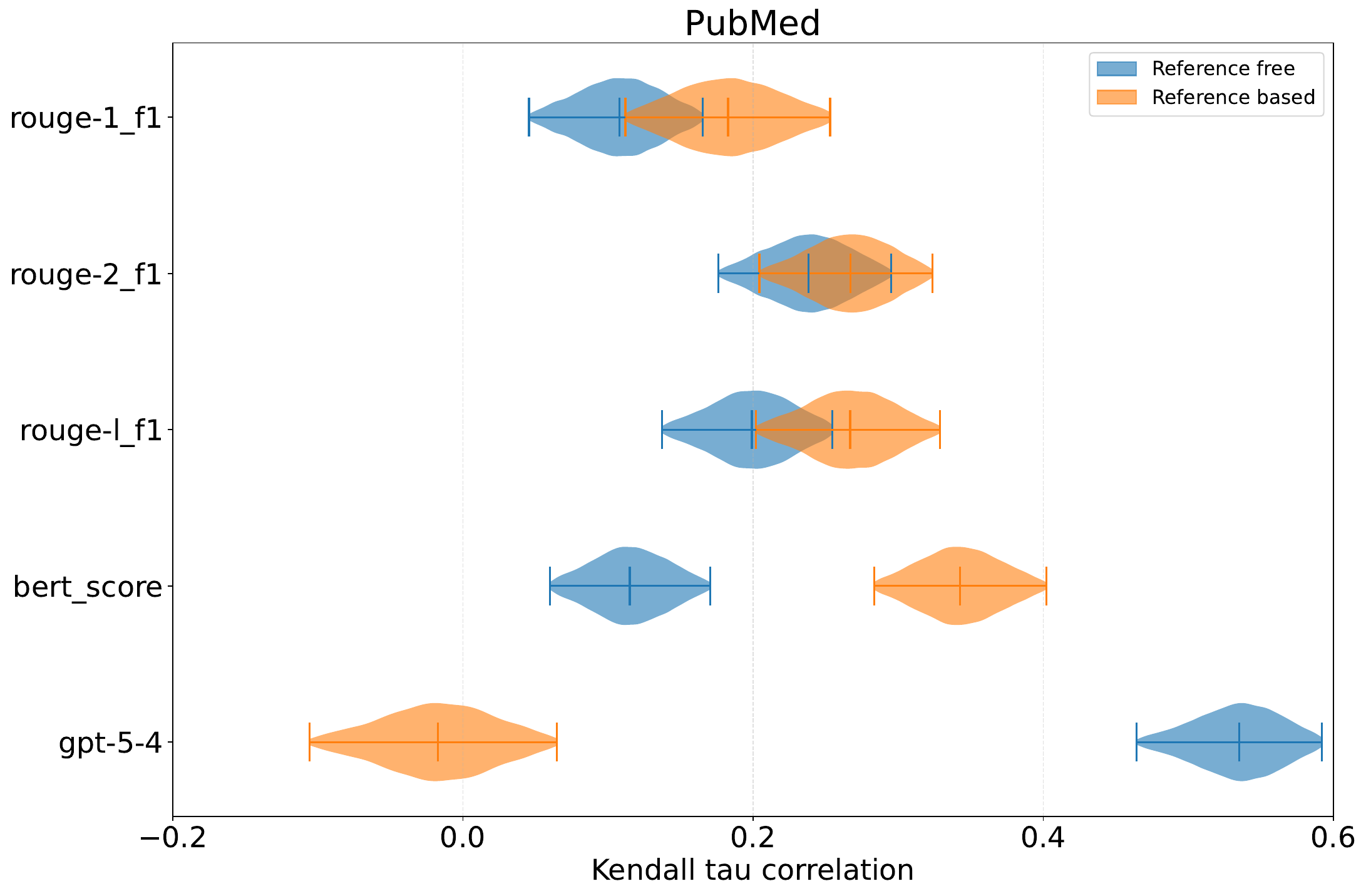}
    \end{subfigure}
    \caption{Violin plot showing 95\% confidence interval of Kendall's $\tau$ correlations of automatic metrics with human judgment on LongEval's PubMed dataset.}
    \label{fig:longeval-correlation}
\end{figure}

Here we provide Figure~\ref{fig:longeval-correlation} showing LongEval results.

\section{Meta Evaluation - Detailed results }\label{app:meta-evaluation-detailed-results}

Table~\ref{tab:appendix_comprehensive_f1} provides statement-level meta-evaluation F1 scores for judges on the pre-extracted statements (Prompt~\ref{prompt:statement-classification-meta-evaluation}, Appendix~\ref{app:prompts}). \uline{Gemini 3.1 Pro dominates, closely followed by Gemma 4 (31B)}. These LLMs outscore human annotators overall, highlighting the significant cognitive load of manual verification. We initially hypothesized that the poor performance of GPT models was due to selection bias, as the adversarial statements were mined using GPT-family models. However, detailed analysis reveals that GPT models also underperform on the non-adversarial subset, largely because they frequently misclassify \textit{True, but wrong citation} and \textit{Unverifiable} labels.

We also analyzed output formatting errors and found that 2.7\% of all statements, created during evaluation of our experiments in Table \ref{tab:results}, are missing the classification field. Nearly all of this (81\% of affected statements) comes from just 26 out of 56~340 generations (0.05\%) in which the judge model entered a degenerate repetition loop. The remaining, much more common failure affects 2~755 generations (4.9\%) where exactly one statement was truncated right before its classification field, without any looping.

\begin{table*}[t]
    \centering
    \small
    \begin{tabular}{llccccc}
        \toprule
        \multirow{2}{*}{\textbf{Evaluator Model}} & \multirow{2}{*}{\textbf{Subset}} & \multicolumn{4}{c}{\textbf{Class-Specific F1 Scores}} & \multirow{2}{*}{\textbf{Macro F1}} \\
        \cmidrule(lr){3-6}
        & & \textbf{True} & \textbf{W. Cite} & \textbf{Unverifiable} & \textbf{False} & \\
        \midrule
        \multirow{3}{*}{\textbf{Human 82}} 
        & All ($N=408$) & 0.74 & 0.72 & 0.46 & 0.57 & 0.62 \\
        & Random ($N=341$) & 0.76 & 0.74 & 0.45 & 0.49 & 0.61 \\
        & Adversarial ($N=67$) & 0.59 & 0.67 & 0.49 & 0.73 & 0.62 \\
        \midrule
        \multirow{3}{*}{\textbf{Human 83}} 
        & All & 0.76 & 0.42 & 0.43 & 0.46 & 0.52 \\
        & Random & 0.76 & 0.40 & 0.41 & 0.47 & 0.51 \\
        & Adversarial & 0.75 & 0.50 & 0.50 & 0.44 & 0.55 \\
        \midrule
        \multirow{3}{*}{\textbf{GPT-5.4}} 
        & All & 0.69 & 0.16 & 0.45 & 0.54 & 0.46 \\
        & Random & 0.72 & 0.16 & 0.51 & 0.53 & 0.48 \\
        & Adversarial & 0.45 & 0.19 & 0.19 & 0.56 & 0.35 \\
        \midrule
        \multirow{3}{*}{\textbf{GPT-5.4-mini}} 
        & All & 0.69 & 0.09 & 0.29 & 0.40 & 0.37 \\
        & Random & 0.71 & 0.09 & 0.35 & 0.38 & 0.38 \\
        & Adversarial & 0.57 & 0.10 & 0.00 & 0.45 & 0.28 \\
        \midrule
        \multirow{3}{*}{\textbf{Gemini 3.1 Pro}} 
        & All & 0.89 & 0.35 & 0.65 & 0.74 & \textbf{0.66} \\
        & Random & 0.90 & 0.30 & 0.73 & 0.76 & \textbf{0.67} \\
        & Adversarial & 0.79 & 0.55 & 0.33 & 0.70 & \textbf{0.59} \\
        \midrule
        \multirow{3}{*}{\textbf{Gemini 3.1 Flash-Lite}} 
        & All & 0.89 & 0.28 & 0.32 & 0.60 & 0.52 \\
        & Random & 0.91 & 0.29 & 0.36 & 0.61 & 0.54 \\
        & Adversarial & 0.79 & 0.27 & 0.17 & 0.58 & 0.45 \\
        \midrule
        \multirow{3}{*}{\textbf{Llama 3.3}} 
        & All & 0.82 & 0.12 & 0.17 & 0.36 & 0.37 \\
        & Random & 0.85 & 0.15 & 0.13 & 0.33 & 0.36 \\
        & Adversarial & 0.65 & 0.00 & 0.27 & 0.43 & 0.34 \\
        \midrule
        \multirow{3}{*}{\textbf{Gemma 4 (31B) Reasoning}} 
        & All & 0.88 & 0.32 & 0.65 & 0.70 & 0.64 \\
        & Random & 0.89 & 0.31 & 0.72 & 0.72 & 0.66 \\
        & Adversarial & 0.76 & 0.38 & 0.38 & 0.65 & 0.54 \\
        \midrule
        \multirow{3}{*}{\textbf{Gemma 4 (31B) Quant. (No Reas.)}} 
        & All & 0.88 & 0.23 & 0.58 & 0.64 & 0.58 \\
        & Random & 0.89 & 0.22 & 0.62 & 0.66 & 0.60 \\
        & Adversarial & 0.76 & 0.25 & 0.38 & 0.59 & 0.49 \\
        \bottomrule
        \bottomrule
    \end{tabular}
    \caption{Detailed performance breakdown of evaluator models across the All, Random, and Adversarial meta-annotated statement subsets. Class-specific F1 scores highlight model calibration on minority classes such as "True, but wrong citation" (W. Cite) and "Unverifiable".}
    \label{tab:appendix_comprehensive_f1}
\end{table*}

\subsection{Note on Coarse Transformation}
When aggregating statement-level classifications into coarse-level faithfulness scores, we observed that a small fraction of samples were associated with no statements. Because this proportion is negligible (e.g., under 0.1\% for the Gemma model), we decided to treat these specific samples as fully faithful.

\section{Annotator Recruitment and Compensation}\label{sec:annotators}
Our human evaluation involved three annotators, all recruited directly from the authors' institutional network (no crowdsourcing platforms were used). The two primary annotators were students in Computer Science, both with completed Bachelor's degrees in Electrical Engineering. The third annotator, responsible for the meta-annotation round was a PhD candidate.

The primary annotators were paid approximately 185 CZK/hour (net wage), roughly 1.6$\times$ the net minimum hourly wage (\textasciitilde113~CZK) close to the national net hourly median (\textasciitilde217 CZK). The PhD candidate participated on a voluntary basis as part of ongoing collaboration on the project. All annotators were informed in advance about the nature of the task.

\section{Licensing and Intended Use}
The dataset contains open-access papers obtained from CORE and SemanticScholar corpora. These corpora contain third-party content and materials, such as open-access works from publicly available sources. In addition to the licenses of those organizations (ODC-By, CC BY-NC), any underlying Third Party Content may be subject to separate license terms by the respective third party owner. We made the best effort to provide identifiers (title, authors, year, DOI, or SemanticScholar ID) of collected papers to allow the user of this dataset to check the license.

The OARelatedWork dataset and the OAPapers corpus are provided solely for research purposes, including but not limited to experimentation with machine learning models and data analysis.

\section{Prompts}\label{app:prompts}
This section provides prompts for used LLMs.

\begin{figure*}[t]
\centering
\refstepcounter{prompt}\label{prompt:statement-extraction-for-human-evaluation}
\begin{tcolorbox}[
    enhanced,
    breakable,
    colback=gray!5!white,
    colframe=gray!60!black,
    title=\textbf{Prompt \theprompt: Statement Extraction Prompt for Human Evaluation},
    fonttitle=\sffamily\bfseries,
    boxrule=0.5mm,
    arc=2mm,
    left=8pt, right=8pt, top=8pt, bottom=8pt
]

\small
\linespread{0.9}
\textbf{\textsf{Role: system}}
\begin{Verbatim}[breaklines,breakanywhere,fontsize=\scriptsize]
You are a scientific research assistant. Your role is to review factual statements within the related work sections of scientific papers and verify their accuracy using only information from the supplied cited papers and the target paper for which the related work was written.
Definition: A factual statement is objective, usually no longer than one sentence, and can be verified as true or false. Do not include subjective opinions or personal experiences.

Format of texts:
- Paragraphs are in <p>...</p> tags.
- Headings are in <h1>...</h1>, <h2>...</h2>, etc. tags.
- Citations are in <cite>ID<sep>paper title<sep>first author</cite> tags. For citations to unknown papers the cite tag contains only <cite>UNK</cite>.
- References to figures and tables are in <ref>figure|table|UNK/ref> tags.
When extracting text leave the tags as is.

Begin with a concise checklist (3-7 bullets) of your approach; keep items conceptual, not implementation-level.
Instructions:
- For each provided statement, determine if it can be verified exclusively using the provided cited papers and the target paper.
- If a statement cannot be verified due to lack of supporting evidence from these sources, mark it as not factual and set the evidence array to empty.
- If there are multiple statements, process each statement in order and return a separate JSON object for each.
- For statements that are malformed or lack citations, set 'is_true' to false and provide an empty 'evidence' array.
Evidence:
- Each item in the 'evidence' array must be an object containing a 'source' field (the ID of the cited paper or 'target_paper') and an 'excerpt' field (a supporting passage from the source).
- The 'evidence' array can have zero or more entries; include all relevant excerpts if multiple sources support your decision.
After processing each statement, briefly validate that your factual assessment and evidence selection are appropriately grounded in the provided sources. If validation fails or is ambiguous, self-correct before finalizing your output.
Output Format:
Return a JSON array. Each element corresponds to a processed input statement and the output follows this schema:
[
{
"statement": "repeat the original statement here, verbatim",
"evidence": [
{
"source": "ID of cited paper or 'target_paper'",
"excerpt": "text excerpt from the source supporting your decision"
}
// ...add more evidence objects as needed
],
"is_true": true | false,
"confidence": confidence score from 0 (completely unsure) to 1 (certain)
}
// ...more elements as needed
]
If no evidence is found or the statement is malformed, set the 'evidence' array empty.
\end{Verbatim}

\textbf{\textsf{Role: user}}
\begin{Verbatim}[breaklines,breakanywhere,fontsize=\scriptsize]
Cited papers:
{% for cp in cited_papers %}
{{cp}}

{% endfor %}

Target paper:
{{target_paper}}

Related work section: 
{{rw}}

Please identify and verify factual statements in the related work section based on the cited papers and target paper.
\end{Verbatim}
\end{tcolorbox}
\end{figure*}

\begin{figure*}[t]
\centering
\refstepcounter{prompt}\label{prompt:statement-classification-meta-evaluation}
\begin{tcolorbox}[
    enhanced,
    breakable,
    colback=gray!5!white,
    colframe=gray!60!black,
    title=\textbf{Prompt \theprompt: Statement Classification Prompt for Meta-Evaluation},
    fonttitle=\sffamily\bfseries,
    boxrule=0.5mm,
    arc=2mm,
    left=8pt, right=8pt, top=8pt, bottom=8pt
]

\small
\linespread{0.9}
\textbf{\textsf{Role: system}}
\begin{Verbatim}[breaklines,breakanywhere,fontsize=\scriptsize]
Role: You are an expert scientific reviewer fact-checking the "Related Work" section of academic papers.

Task: You are given a factual statement from the provided "Related Work" section; verify it against the provided reference material (the rest of the paper and the cited papers). Do not use the Related Work section itself as reference material for verification.

Constraints:
- KNOWLEDGE BOUNDARY: Base your decisions strictly and solely on the provided input text. Do not use external knowledge or training data.

Verification Categories (For your internal processing):
1. "True": The statement is fully supported by the reference material AND the specific citation provided is correct.
2. "True, but wrong citation": The statement is factually supported by the reference material, but the specific citation attached to it does not support it.
3. "Unverifiable": The provided reference material does not contain enough information to prove the statement true or false.
4. "False": The statement is directly contradicted by the reference material.

Instructions:
1. Read the context of the statement in the "Related Work" section to understand its meaning and the specific citation attached to it. The statement is highlighted in the provided "Related Work" section with <statement> tags.
2. Search for evidence in the reference material that can be used to classify the statement into one of the four Verification Categories. The evidence must be directly quoted text from the reference material that supports or contradicts the statement.
3. Write a rationale (maximum two sentences) explaining your final decision.
4. Make a final classification into one of the four Verification Categories.

Output Format:
Output the following information in a structured format (e.g., JSON):
{
  "evidence": [
    "direct quote from the reference material that supports or contradicts the statement"
  ],
  "rationale": "a brief explanation of the reasoning behind the classification",
  "classification": "one of the four Verification Categories (True, True but wrong citation, Unverifiable, False)"
}
\end{Verbatim}

\textbf{\textsf{Role: user}}
\begin{Verbatim}[breaklines,breakanywhere,fontsize=\scriptsize]
Please review the following factual statement from the "Related Work" section and evaluate it against the provided reference materials.
            
<reference_material_rest_of_paper>
{{ txt_target_paper | truncate(tokenizer_name="casperhansen/llama-3.3-70b-instruct-awq", number_of_tokens=30000, direction="right", backend="transformers") }}
</reference_material_rest_of_paper>
            
<reference_material_citations>
{{ txt_cited_papers | truncate(tokenizer_name="casperhansen/llama-3.3-70b-instruct-awq", number_of_tokens=90000, direction="right", backend="transformers") }}
</reference_material_citations>

<statement>
{{ text }}
</statement>

<related_work>
{{ txt_rw[:start] }}<statement>{{ txt_rw[start:end] }}</statement>{{ txt_rw[end:] }}
</related_work>
\end{Verbatim}
\end{tcolorbox}
\end{figure*}

\begin{figure*}[t]
\centering
\refstepcounter{prompt}\label{prompt:the-judge-prompt}
\begin{tcolorbox}[
    enhanced,
    breakable,
    colback=gray!5!white,
    colframe=gray!60!black,
    title=\textbf{Prompt \theprompt: The Judge Prompt},
    fonttitle=\sffamily\bfseries,
    boxrule=0.5mm,
    arc=2mm,
    left=8pt, right=8pt, top=8pt, bottom=8pt
]

\small
\linespread{0.9}
\textbf{\textsf{Role: system}}
\begin{Verbatim}[breaklines,breakanywhere,fontsize=\scriptsize]
Role: You are an expert scientific reviewer fact-checking the "Related Work" section of academic papers.

Task: Extract all factual statements from the provided "Related Work" section, verify them against the provided reference material (the rest of the paper and cited papers).

Constraints:
- KNOWLEDGE BOUNDARY: Base your decisions strictly and solely on the provided input text. Do not use external knowledge or training data.

Definitions:
- Factual Statement: An objective claim (typically one sentence or less) devoid of subjective opinion, capable of being verified against the reference material.

Verification Categories (For your internal processing):
1. "True": The statement is fully supported by the reference material AND the specific citation provided is correct.
2. "True, but wrong citation": The statement is factually supported by the reference material, but the specific citation attached to it does not support it.
3. "Unverifiable": The provided reference material does not contain enough information to prove the statement true or false.
4. "False": The statement is directly contradicted by the reference material.

Instructions:
1. Systematically identify every factual statement in the "Related Work" section.
2. Search for evidence in the reference material that can be used to classify each statement into one of the four Verification Categories. The evidence can only be directly quoted text from the reference material that supports or contradicts the statement.
3. Write a rationale (maximum two sentences) explaining your final decision.
4. Make a final classification for each statement into one of the four Verification Categories.

Output Format:
For each factual statement, output the following information in a structured format (e.g., JSON):
{
  "statements": [
    {
      "text": "exact text of the factual statement extracted from the Related Work section",
      "evidence": [
        "direct quote from the reference material that supports or contradicts the statement"
      ],
      "rationale": "a brief explanation of the reasoning behind the classification",
      "classification": "one of the four Verification Categories (True, True but wrong citation, Unverifiable, False)"
    },
    ...
  ]
}
\end{Verbatim}

\textbf{\textsf{Role: user}}
\begin{Verbatim}[breaklines,breakanywhere,fontsize=\scriptsize]
Please review the following "Related Work" section and evaluate its factual statements against the provided reference materials.

<reference_material_rest_of_paper>
{{ target_paper }}
</reference_material_rest_of_paper>

<reference_material_citations>
{{ cited_papers }}
</reference_material_citations>

<related_work>
{{ summary }}
</related_work>

\end{Verbatim}
\end{tcolorbox}
\end{figure*}

\begin{figure*}[t]
\centering
\refstepcounter{prompt}\label{prompt:llm-baselines-part-1}
\begin{tcolorbox}[
    enhanced,
    breakable,
    colback=gray!5!white,
    colframe=gray!60!black,
    title=\textbf{Prompt \theprompt: LLM baselines prompt when all content is used (PART 1)},
    fonttitle=\sffamily\bfseries,
    boxrule=0.5mm,
    arc=2mm,
    left=8pt, right=8pt, top=8pt, bottom=8pt
]

\small
\linespread{0.9}
\textbf{\textsf{Role: system}}
\begin{Verbatim}[breaklines,breakanywhere,fontsize=\scriptsize]
Write the related work section for the paper described bellow and given references.
          
Format instructions:
  Write paragraphs on separate lines.
  Use in-text citations for references using following format: <cite>ID<sep>TITLE<sep>FIRST_AUTHOR</cite>.
  {% if structured %}Use subsections for different topics. Every subsection has a title on a separate line with a # prefix. For the first subsection, use ##, for the second, ###, etc.{% else %}Do not use subsections.{% endif %}
  Write just the content of the related work section. This means that you are not supposed to write the title of the section or the references section.


An example of related work section:

{% if structured %}
## 2.1 Load balancing
Many research have been done to address the issue of load balancing in distributed systems. They are typically classified into static and dynamic ones <cite>30331579<sep>Comparison of dynamic and static load-balancing strategies in heterogeneous distributed systems<sep>Y Zhang</cite>. Static policies determine job assignments based on information about the average behavior of the system. They are simple, efficient, and easy to implement, but cannot adapt to the changing workload in the system. To respond to the fluctuation of workload over time, dynamic load balancing policies are proposed. Ref. <cite>18652554<sep>A dynamic load-balancing policy with a central job dispatcher (LBC)<sep>H C Lin</cite>, for example, gives a policy with a central job dispatcher, which periodically collects load information at system nodes and makes decisions for job transformation. They are more complex than static policies, but exhibit better performance. These load balancing policies view system nodes as a collection of common resources for arriving jobs. This makes them inapplicable to the scenario we are concerned about, in which a query should be processed by a particular server where the queried data are located. Refs. <cite>10822818<sep>Data distribution algorithms for load balanced fault-tolerant web access<sep>B Narendran</cite><cite>5163968<sep>Approximation algorithms for data distribution with load balancing of web servers<sep>L C Chen</cite><cite>5700028<sep>Approximate algorithms for document placement in distributed web servers<sep>S Tse</cite> address the issue of data allocation; however they are typically designed to allocate web documents among a cluster of web servers, where the document size and HTTP connection numbers are main concerns.
A more relative work is data distribution in parallel storage systems. In these systems, horizontal data partitioning has been commonly used to distribute data among system nodes. There are typically three strategies for horizontal partitioning: round-robin, hash, and value-range partitioning. These partitioning strategies each has advantages and disadvantages, but they do not take into consideration the access pattern on data, which would be a problem when the pattern is not a uniform distribution. Dynamic data reallocation needs to effectively adapt to changes of access patterns. When access frequencies of data items lead to unbalanced workload on system nodes, data migration should be performed to balance workload. The simple data skew handling method in <cite>18421049<sep>Fat-Btree: An update-conscious parallel directory structure<sep>H Yokota</cite> balances the storage space for data, but does not guarantee balanced data access load across system nodes. Similarly most access load skew handling methods do not consider data balancing <cite>4657232<sep>Towards self-tuning data placement in parallel database systems<sep>M L Lee</cite><cite>111471344<sep>A fast convergence technique for online heat-balancing of Btree indexed database over shared-nothing parallel systems<sep>H Feelifl</cite>. In <cite>12316292<sep>Adaptive lapped declustering: A highly available data-placement method balancing access load and space utilization<sep>A Watanabe</cite>, a data-placement method was proposed to balance both access load and storage space of data, but it is designed for parallel storage system, and focuses on achieving availability and scalability for parallel storage configuration.
## 2.2 Market mechanism
During the past few years, market mechanism has gained much interest in both the economics and computer science literature. Market-based techniques <cite>3303225<sep>Characterizing effective auction mechanisms: insights from the 2007 tac market design competition<sep>J Niu</cite> are applying themselves to resource allocation, often through the creation of artificial market. This is due to the fact that market mechanisms, when well designed, can achieve desired economic outcomes like high allocated efficiency. Auctions are a subclass of market mechanisms that have received particular attention. Double auction <cite>117078532<sep>The double auction institution: A survey<sep>D Friedman</cite>, on which our method is based, is one type of auction where both buyers and sellers can make offers. There is a wide range of literature published on double auctions (e.g., <cite>1372607<sep>Market power and efficiency in a computational electricity market with discriminatory double-auction pricing<sep>J Nicolaisen</cite><cite>124078139<sep>Pricing in agent economies using multi-agent Q-learning<sep>G Tesauro</cite>).
Connecting load balancing to the market mechanism in the distributed environment is a new direction of research. In our approach, we will show how to integrate the ideas of the artificial market into solving load balancing challenges in the cloud-based environment.
{% else %}

\end{Verbatim}

\end{tcolorbox}
\end{figure*}

\begin{figure*}[t]
\centering
\refstepcounter{prompt}\label{prompt:llm-baselines-part-2}
\begin{tcolorbox}[
    enhanced,
    breakable,
    colback=gray!5!white,
    colframe=gray!60!black,
    title=\textbf{Prompt \theprompt: LLM baselines prompt when all content is used (PART 2)},
    fonttitle=\sffamily\bfseries,
    boxrule=0.5mm,
    arc=2mm,
    left=8pt, right=8pt, top=8pt, bottom=8pt
]

\small
\linespread{0.9}

\begin{Verbatim}[breaklines,breakanywhere,fontsize=\scriptsize]
Recent years have witnessed the huge growth of the Internet, and the emergence of new applications that require guaranteed service levels. Real-time consumer applications, such as streaming live video and online gaming have introduced new traffic characteristics and imposed new requirements on network performance and resource availability. So far, the Internet has offered only best-effort services. All traffic is delivered in the best possible way with no guarantees given to any type of traffic. However, today there are more applications that demand service guarantees in order to function properly. Supporting these new types of applications requires more sophisticated routing mechanisms for path selection and for better network resource utilization. The concept of Quality of Service (QoS) has been recently addressed extensively in the literature as a means of providing the required service level for demanding applications <cite>13100834<sep>Survey of qos routing<sep>P Paul</cite>.
QoS routing is a major component of the QoS paradigm; it consists of a set of routing related mechanisms that guarantee to provide the service level required by traffic flows. In QoS routing, paths for flows are selected based upon knowledge of the resource availability, referred to as either link-state information or network-state information, held by the network nodes and the QoS requirements of the flows.
QoS routing selects a path that has sufficient resources to accommodate the QoS requirement of a given flow <cite>5747745<sep>A Framework for QoS-based Routing in the Internet<sep>R N E Crawley</cite>. In QoS routing, some knowledge regarding the global link state is required by each network node in order to take routing decisions.
This knowledge can be obtained through a periodic exchange of the network state among the network nodes. Using link-state updates, each node constructs a global view of the network state and selects the best path for a flow based on this global view of the network. Global QoS routing performs well when each source node has a reasonably accurate view of the entire network state. However, the network state changes with each flow arrival or departure and maintaining an accurate network state becomes impractical, due to the rapid flow dynamics. Moreover, the exchange of link states among nodes introduces prohibitive communication and processing overheads. As a result of inaccurate state information, routing may suffer degraded performance and instability <cite>UNK<sep>Quality of Service in IP networks<sep>G Armitge</cite><cite>9587075<sep>An Overview of Quality-of-Service Routing for the Next Generation HighSpeed Networks: Problems and Solutions<sep>S Chen</cite>. Consequently, an alternative routing approach that can circumvent the problems associated with global state information is sought.
{% endif %}
\end{Verbatim}

\textbf{\textsf{Role: user}}
\begin{Verbatim}[breaklines,breakanywhere,fontsize=\scriptsize]
Write the related work section for the paper described bellow and given references.
  
The related work section should have around {{summary_length_words}} words. 

Description of the paper:{% set truncated_queries = queries | truncate_group(max_length=22000) %}
Title:
{{truncated_queries[0]}}

Abstract:
{{truncated_queries[1]}}

Rest of the paper:
{{truncated_queries[2]}}
\end{Verbatim}

\textbf{\textsf{Role: assistant}}
\begin{Verbatim}[breaklines,breakanywhere,fontsize=\scriptsize]
Please provide references that should be used in the related work section.
\end{Verbatim}

\textbf{\textsf{Role: user}}
\begin{Verbatim}[breaklines,breakanywhere,fontsize=\scriptsize]
Sure, here are the references:
  
{% for paper in inputs | truncate_group(max_length=80000) %}
{{paper}}

{% endfor %}

Use them to write the related work section.
\end{Verbatim}
\end{tcolorbox}
\end{figure*}

\end{document}